\documentclass{article}

\usepackage{microtype}
\usepackage{graphicx}
\usepackage{subfigure}
\usepackage{booktabs} % for professional tables

\usepackage{hyperref}

\usepackage[accepted]{icml2024}

\usepackage{amsmath}
\usepackage{amssymb}
\usepackage{mathtools}
\usepackage{amsthm}

\usepackage{bm}
\usepackage{multirow}
\usepackage{pifont}

\usepackage[capitalize,noabbrev]{cleveref}

\theoremstyle{plain}
\newtheorem{theorem}{Theorem}[section]
\newtheorem{proposition}[theorem]{Proposition}

\newtheorem{corollary}[theorem]{Corollary}
\theoremstyle{definition}
\newtheorem{definition}[theorem]{Definition}

\theoremstyle{remark}

\usepackage[textsize=tiny]{todonotes}

\icmltitlerunning{Resisting Stochastic Risks in Diffusion Planners with the Trajectory Aggregation Tree}

\begin{document}

\twocolumn[
\icmltitle{Resisting Stochastic Risks in Diffusion Planners with \\ the Trajectory Aggregation Tree}



\icmlsetsymbol{equal}{*}

\begin{icmlauthorlist}
\icmlauthor{Lang Feng}{zju}
\icmlauthor{Pengjie Gu}{ntu}
\icmlauthor{Bo An}{ntu,sky}
\icmlauthor{Gang Pan}{zju,keylabzju}
\end{icmlauthorlist}

\icmlaffiliation{ntu}{Nanyang Technological University, Singapore}
\icmlaffiliation{sky}{Skywork AI, Singapore}
\icmlaffiliation{zju}{Zhejiang University, China}
\icmlaffiliation{keylabzju}{State Key Laboratory of Brain-Machine Intelligence, China}

% \icmlcorrespondingauthor{Bo An}{boan@ntu.edu.sg}
\icmlcorrespondingauthor{Gang Pan}{gpan@zju.edu.cn}

\icmlkeywords{Machine Learning, ICML}

\vskip 0.3in
]



\printAffiliationsAndNotice{}  % leave blank if no need to mention equal contribution

\begin{abstract}
Diffusion planners have shown promise in handling long-horizon and sparse-reward tasks due to the non-autoregressive plan generation. However, their inherent stochastic risk of generating infeasible trajectories presents significant challenges to their reliability and stability. We introduce a novel approach, the Trajectory Aggregation Tree (TAT), to address this issue in diffusion planners. Compared to prior methods that rely solely on raw trajectory predictions, TAT aggregates information from both historical and current trajectories, forming a dynamic tree-like structure. Each trajectory is conceptualized as a branch and individual states as nodes. As the structure evolves with the integration of new trajectories, unreliable states are marginalized, and the most impactful nodes are prioritized for decision-making. TAT can be deployed without modifying the original training and sampling pipelines of diffusion planners, making it a training-free, ready-to-deploy solution. We provide both theoretical analysis and empirical evidence to support TAT's effectiveness. Our results highlight its remarkable ability to resist the risk from unreliable trajectories, guarantee the performance boosting of diffusion planners in 100\% of tasks, and exhibit an appreciable tolerance margin for sample quality, thereby enabling planning with a more than $3\times$ acceleration.
\end{abstract}

\section{Introduction}
Planning is a crucial aspect of decision-making and has led to many successful outcomes, like simulated robot control~\cite{tassa2012synthesis,varlamov2024new} and board games~\cite{silver2016mastering,silver2017mastering}. Conventionally, planning relies on the complete model of the environment that provides the necessary information about action effects and outcome probabilities.
In instances where such environment's dynamics are not available, model learning and function approximation techniques are usually employed~\cite{deisenroth2011pilco,levine2014learning,hafner2019learning,schrittwieser2020mastering}, like model-based reinforcement learning~\cite{sutton2018reinforcement}.

Recent advances have focused on leveraging diffusion models~\cite{ho2020denoising}, known for their ability to generate high-quality samples~\cite{luo2021diffusion,li2022diffusion,nichol2022glide}, to create diffusion-based planners~\cite{janner2022planning,ajay2023is,liang2023adaptdiffuser}. 
These planners approach sequential decision optimization as a data-driven trajectory optimization problem and simultaneously predict all timesteps of a trajectory. This innovative paradigm achieves non-autoregressive plan generation and aligns model learning with planning, thereby avoiding the compounding rollout errors~\cite{asadi2018lipschitz} and adversarial plans~\cite{talvitie2014model,ke2018modeling}. Impressively, diffusion planners have demonstrated remarkable abilities to handle challenges like long-horizon, sparse-reward, as well as offline control tasks~\cite{fu2020d4rl}.

Despite these advantages, diffusion planners are unable to guarantee the generation of reliable and feasible plans due to the inherent \emph{stochasticity} of diffusion models. Unlike deterministic models that yield the same output for a given input, the outputs of diffusion models are probabilistic. This characteristic means that while diffusion models may often produce high-quality samples, there is always a risk of generating unreliable samples, which are referred to as ``\emph{artifacts}'' in vision task~\cite{bau2019gan,shen2020interpreting}. 

Current diffusion planners directly base their decisions on trajectories produced by diffusion models ~\cite{janner2022planning,ajay2023is,liang2023adaptdiffuser}. However, these planners often presuppose the high quality of the generated trajectories, neglecting the stochastic risks inherent in the process. This neglect can result in decision-making reliant on unreliable trajectories, leading to potentially ineffective or harmful actions for the intended task. Consequently, the stability and reliability of these diffusion planners are compromised, particularly in safety-critical applications \cite{lee2023refining}. Moreover, to enhance the generation of effective trajectories, these planners necessitate extra denoising steps. Such steps, however, lead to considerable delays in the decision-making process and introduce a critical limitation for scenarios that demand real-time responses, such as autonomous vehicle navigation~\cite{auto_vehicle}. 

In this study, we explore a novel approach inspired by the ``wisdom of the crowd" concept~\cite{sagi2018ensemble}, which posits that the collective opinion of a group often exceeds the accuracy of an individual's judgment. Our proposal, the Trajectory Aggregation Tree (TAT), aims to enhance decision-making in diffusion planners by resisting stochastic risks. Contrasting with previous methods that rely solely on raw trajectory predictions, TAT integrates and analyzes collective data from both historical and current trajectories.
TAT functions as a dynamic tree-like structure. Each trajectory is conceptualized as a branch, with individual states forming the nodes. This structure evolves as new trajectories are integrated, akin to a tree growing branches. The impact of each trajectory on the TAT's overall structure is variable, with more trajectories exerting a more substantial influence on specific nodes. This mechanism prioritizes vital data while filtering out less reliable inputs.  Ultimately, TAT makes its decisions based on the most impactful nodes, thereby enhancing the robustness and reliability.
    
Our proposed TAT has several appealing advantages: \textbf{(1)}~Our study provides a theoretical justification to prove TAT's effectiveness in mitigating the impact of artifacts via trajectory aggregation. In our analysis, TAT-reinforced diffusion planners always outperform the original ones. \textbf{(2)}~TAT exhibits an appreciable tolerance margin for artifact generation. This feature allows diffusion planners to sacrifice the generation quality, such as reducing denoising steps, to speed up planning without a noticeable performance decline. \textbf{(3)}~TAT functions as a training-free and ready-to-deploy solution for existing diffusion planners; there is no need for fine-tuning and re-training of the original model. \textbf{(4)}~In our empirical evaluations, we develop TAT on various diffusion planners, subjecting them to a diverse set of decision-making tasks. Our results demonstrate its impressive ability to filter out stochastic artifacts, consistently improve the performance of diffusion planners in all tasks, and enable planning speeds more than threefold faster. 
% Source code and materials are available at \url{https://github.com/langfengQ/tree-diffusion-planner}.

\section{Related Work}
Diffusion probabilistic models~\cite{sohl2015deep,ho2020denoising} have proven their highly-expressive generative capabilities in various domains, like computer vision~\cite{lugmayr2022repaint,brempong2022denoising,fang2024alleviating}, natural language processing~\cite{austin2021structured,li2022diffusion}, and temporal data modeling~\cite{tashiro2021csdi,kong2021diffwave,wang2024diffmdd}. In recent developments, diffusion models have found applications in reinforcement learning (RL), with a particular focus on offline settings. These applications involves the use of diffusion models as planners~\cite{janner2022planning,ajay2023is}, policies~\cite{wang2023diffusion,chen2023offline}, and for data augmentation~\cite{chen2023genaug}.

Regarding planners, one of the representative works is the Diffuser~\cite{janner2022planning}, which plans by iteratively denoising trajectories using a diffusion model. It combines trajectory optimization with modeling, making sampling from the model and planning with it nearly identical. The Diffuser has demonstrated the long-horizon planning and test-time flexibility in offline control tasks. Building upon this framework, recent works have proposed several improvements. For example, Decision Diffuser~\cite{ajay2023is} incorporates classifier-free guidance, therefore avoiding the need for a separate reward function. AdaptDiffuser~\cite{liang2023adaptdiffuser} uses an evolutionary planning method to self-evolve the diffusion model and generate rich synthetic expert data for goal-conditioned tasks. \citet{lee2023refining} introduced the restoration gap to enhance the quality of trajectories, which requires training an additional gap predictor and using its gradient to refine the sampling of the diffusion model. Additionally, diffusion planners have been applied to multi-task RL~\cite{he2023diffusion,ni2023metadiffuser}, hierarchical RL~\cite{li2023hierarchical,chen2024simple}, and multi-agent RL~\cite{zhu2023madiff}. Despite these advances, the probabilistic denoising process of diffusion models leads to the stochastic risk of yielding impractical and unreliable plans. This unpredictability introduces considerable challenges to ensuring the stability and reliability of these planners.

\section{Background}
\label{sec:background}

\paragraph{Problem Setting.}
Consider a discounted Markov decision process (MDP), defined by the tuple $(\mathcal{S},\mathcal{A},P,r,\gamma)$. Here, $\mathcal{S}$ and $\mathcal{A}$ are the state space and action space respectively. The transition probability function $P:\mathcal{S}\times \mathcal{A} \times \mathcal{S} \rightarrow [0, 1]$ determines the likelihood of moving from one state to another given a specific action. $r: \mathcal{S}\times \mathcal{A}\rightarrow \mathbb{R}$ is the reward function. $\gamma \in [0, 1)$ is the discount factor. Within each given state $\textbf{s}_t\in\mathcal{S}$, the agent acts with action $\textbf{a}_t\in\mathcal{A}$. This action leads to a transition to a new state: $\textbf{s}_{t+1}=P(\textbf{s}_t,\textbf{a}_t)$, and the agent receives a reward $r(\textbf{s}_t,\textbf{a}_t)$. 

In the sequence modeling, the goal of trajectory optimization is to identify an optimal sequence of actions $\textbf{a}^{\ast}_{0:T}$ that maximizes the expected return $\mathcal{J}(\bm{\tau})$ with respect to trajectory $\bm{\tau}=(\textbf{s}_0,\textbf{a}_0,\textbf{s}_1,\textbf{a}_1,\dots,\textbf{s}_T,\textbf{a}_T)$:
\begin{equation}
\textbf{a}^{\ast}_{0:T}=\mathop{\arg\max}\limits_{\textbf{a}_{0:T}}\mathcal{J}(\bm{\tau})=\mathop{\arg\max}\limits_{\textbf{a}_{0:T}}\sum_{t=0}^T\gamma^tr(\textbf{s}_t,\textbf{a}_t),
\end{equation}
where $t$ denotes the planning step and $T$ denotes the planning horizon.

\paragraph{Diffusion Model.}
Synthesizing high-quality plans and behaviors entails learning a model distribution $p_{\theta}(\bm{\tau})$ that effectively approximates the ground-truth distribution of the trajectories $q(\bm{\tau})$. To this end, diffusion probabilistic models~\cite{ho2020denoising} typically factorize the data distribution $p_{\theta}(\bm{\tau})$ as a Markov chain of Gaussian transitions:
\begin{equation}
   p_{\theta}(\bm{\tau}_0)=\int p_{\theta}(\bm{\tau}_{K})\prod_k^K p_{\theta}(\bm{\tau}_{k-1}|\bm{\tau}_{k}) d\bm{\tau}_{1:K},
\end{equation}
where $\bm{\tau}_{0}$ denotes the original (noiseless) data, $\bm{\tau}_{1},\dots,\bm{\tau}_{K}$ denote the latent variables with the same dimensionality of $\bm{\tau}_{0}$, $p_{\theta}(\bm{\tau}_K)=\mathcal{N}(\bm{\tau}_K;\bm{0},\textbf{I})$ is the Gaussian prior, and $p_{\theta}(\bm{\tau}_{k-1}|\bm{\tau}_k)$ is the trainable reverse denoising process defined by 
\begin{equation}
p_{\theta}(\bm{\tau}_{k-1}|\bm{\tau}_k)=\mathcal{N}(\bm{\tau}_{k-1};\mu_{\theta}(\bm{\tau}_{k},k),\Sigma_{\theta}(\bm{\tau}_{k},k)).
\end{equation}
Diffusion model predefines a forward noising process $q(\bm{\tau}_k|\bm{\tau}_{k-1})=\mathcal{N}(\bm{\tau}_k;\sqrt{1-\beta_k}\bm{\tau}_{k-1},\beta_k\textbf{I})$, which progressively adding Gaussian noise to the data according to the variance schedule $\beta_k$. The trajectory generation involves an iterative denoising procedure $p_{\theta}(\bm{\tau}_{k-1}|\bm{\tau}_k)$, starting from a Gaussian noise $\bm{\tau}_K$.

However, a tractable variational lower-bound exists on $\mathbb{E}_{q(\bm{\tau}_0)}[\log p_{\theta}(\bm{\tau}_0)]$. A better scheme arises from fixing $\Sigma_{\theta}(\bm{\tau}_k,k)$ and optimizing a surrogate loss~\cite{ho2020denoising}:
\begin{equation}
   \mathcal{L}^{\text{simple}}(\theta)=\mathbb{E}_{k\sim\mathcal{U},\bm{\tau}_0\sim q, \epsilon\sim\mathcal{N}(\bm{0},\textbf{I})}\left[||\epsilon-\epsilon_{\theta}(\bm{\tau}_k,k)||^{2}\right],
\end{equation}
where $\mathcal{U}$ is a uniform distribution between $1$ and $K$. The nosie-predictor model $\epsilon_{\theta}(\bm{\tau}_k,k)$ aims to estimates the added noise, from which $\mu_{\theta}(\bm{\tau}_{k},k)$ can be readily derived.

To sample trajectories meeting certain constraints, it is necessary to formulate the problem as a conditional distribution $p_{\theta}(\bm{\tau}|y(\bm{\tau}))$, where $y(\bm{\tau})$ is some specific condition on trajectory sample $\bm{\tau}$. There are two common choices to do so: classifier guidance~\cite{dhariwal2021diffusion} and classifier-free guidance~\cite{ho2021classifierfree}. See Appendix~\ref{sec:appendix_guided_diffusion}.

\paragraph{Planning with Diffusion.}
Given a well-trained diffusion model, planning is based on synthesizing a trajectory $\bm{\tau}_0$ through an iterative denoising process that progresses from $\bm{\tau}_K$ to $\bm{\tau}_0$. The trajectory is mathematically represented as:
\begin{equation}
   \bm{\tau}_k = \big\{\bm{x}_0,\bm{x}_1,\bm{x}_{2},\dots,\bm{x}_{T}\big\}_k,
\end{equation}
where $k$ indicates the denoising sample step. The trajectory element, $\bm{x}_t$, manifests in two distinct forms, indicative of differing decision-making strategies: (1) The first is state-centric form, with $\bm{x}_t$ being solely the state, denoted as $\bm{x}_t=\textbf{s}_t$~\cite{ajay2023is}. In this format, action generation is computed using an inverse dynamics model~\cite{agrawal2016learning,pathak2018zero}, based on consecutive states: $\textbf{a}_t=f_{\phi}(\textbf{s}_t,\textbf{s}_{t+1})$.
(2) The second is state-action form, integrating state and preceding action: $\bm{x}_t=(\textbf{s}_t,\textbf{a}_{t-1})$~\cite{janner2022planning,liang2023adaptdiffuser}, where $\textbf{a}_{-1}$ in $\bm{x}_0$ indicates an absence of action. This form enables direct extraction of actions for decision-making. More details of diffusion planners are provided in Appendix~\ref{sec:appendix_diffuser_details}.

In these planning strategies, the quality of trajectory $\bm{\tau}$ (short for $\bm{\tau}_0$) is essential: trajectories of low quality can greatly disrupt action generation and decision-making. Unfortunately, the stochasticity of diffusions often leads to the generation of unreliable trajectories, exacerbating the severity of this issue within prior planning strategies.
Additionally, to adapt to changing environments, the planner strategically creates a new trajectory for planning at each step, leaving the old one behind, even though it may contain valuable information for future decisions.

\section{Trajectory Aggregation Tree}
\begin{figure}[t]
   \vskip 0.1in
   \begin{center}
   \centerline{\includegraphics[width=0.98\columnwidth]{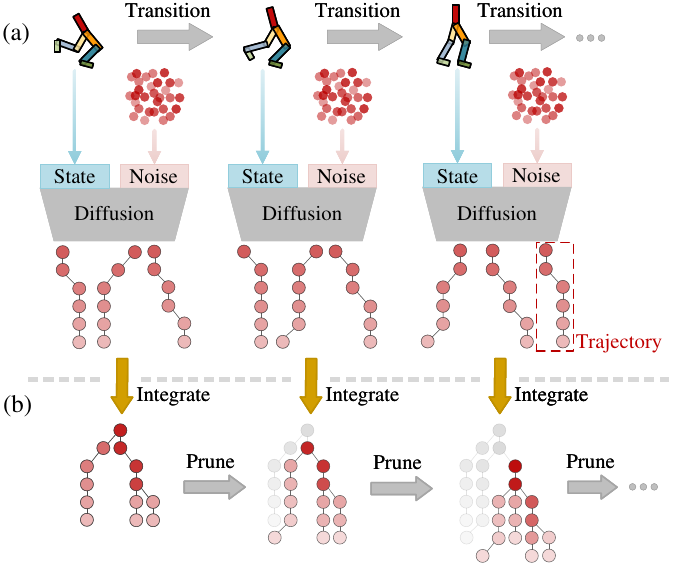}}
   \caption{Overview of TAT. (a) Diffusion pipeline. Given the environment state, TAT utilizes the original diffusion planner to sample trajectories. (b) TAT pipeline. The tree incorporates past and current trajectories to construct a comprehensive experience of future states. The darker the red color, the higher the weight the node has. The tree grows dynamically and keeps in sync with the environment by pruning branches (the light transparent gray part).}
   \label{fig:tree_overview}
   \end{center}
   \vskip -0.15in
\end{figure}
In response to the stochastic risks in diffusion planners, we introduce the Trajectory Aggregation Tree (TAT) method.
TAT innovatively utilizes data from both current and previous steps, akin to ensemble learning techniques that harness collective wisdom to mitigate individual uncertainties. 
Figure~\ref{fig:tree_overview} provides an overview of TAT and illustrates its key aspects. It conceptualizes each trajectory $\bm{\tau} = \{\bm{x}_0,\dots,\bm{x}_T\}$ as a branch and introduces a tree structure to \textbf{merge} trajectories with similar states, while \textbf{branching out} for divergent trajectories. This procedure is \textbf{executed efficiently}, as TAT operates without necessitating new models, extra trainable parameters, or additional generation steps. In Figure~\ref{fig:tree_overview}, we utilize the original pipeline of diffusion planners to generate trajectories, treating TAT as a downstream process. Moreover, TAT grows \textbf{dynamically} as the diffusion cycle repeats and keeps up-to-date with the environment, facilitating real-time planning. With each timestep, it seamlessly incorporates new trajectories from the diffusion pipeline, enriching its structure with additional data. TAT evaluates the \textbf{cumulative effect} of all trajectories on each node by assigning weights. This strategy emphasizes fundamental nodes assigned with the highest cumulative weights while pruning substandard individual nodes, effectively mitigating the impact of stochastic artifacts.

In the following sections, we will delve into the node structure of TAT, outline the process of constructing TAT and its application in planning, and provide a theoretical analysis of TAT's effectiveness in mitigating risks associated with unreliable data generation.

\subsection{Node Structure}
We next establish the node structure of the TAT. Our objective is to enable the aggregation of analogous information across multiple trajectories while weighting decision-related data. To this end, we define each node within the tree by the expression $e_{t,i}$. Each node is represented by the tuple:
\begin{equation}
e_{t,i} = \big\{X(e_{t,i}), V(e_{t,i}), \bm{x}(e_{t,i})\big\},
\end{equation}
where $t$ indicates the node's depth, corresponding to the environmental step, and $i=0,1,\ldots$ indexes the nodes under the same parent. The edges between nodes symbolize temporal relationships. Node $e_{t,i}$ comprises a set of statistics: 
\begin{itemize}
\item $X(e_{t,i})$ contains a set of \textbf{similar trajectory elements} $\bm{x}_t$ from multiple trajectories. $\bm{x}_t$ is representative of either state $\textbf{s}_t$ or state-action pairs $(\textbf{s}_t, \textbf{a}_{t-1})$, based on the planning strategies outlined in Section~\ref{sec:background}.
\item $V(e_{t,i})$ contains the \textbf{weights} assigned to node $e_{t,i}$ from these trajectory elements, reflecting its visitation frequency and significance.
\item $\bm{x}(e_{t,i})$, the \textbf{node state}, distills the joint feature from the elements within $X(e_{t,i})$, serving as a canonical representation of the node’s aggregated data.
\end{itemize}

Further elaboration on these statistics will be provided in the subsequent sections, detailing their operational roles.

\subsection{Planning with TAT}
\begin{figure*}[t]
   \vskip 0.1in
   \begin{center}
   \centerline{\includegraphics[width=0.95\textwidth]{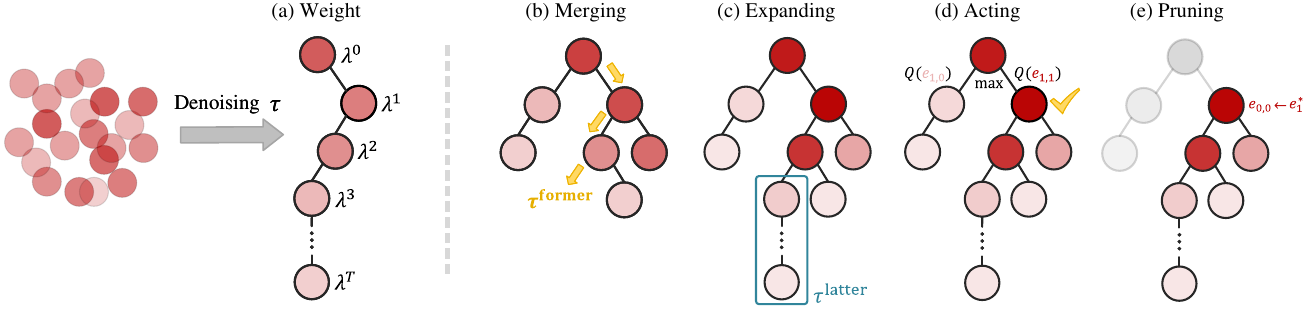}}
   \caption{Planning with TAT. The darker the red color, the higher the weight the node has. (a) Weight allocation to each state in the sampled trajectory $\bm{\tau}$. (b) Merging procedure starting from the root node and sequentially traversing the states in $\bm{\tau}$ to assign weights to the tree nodes. (c) Expanding the tree with the sub-trajectory $\bm{\tau}^{\rm latter}$ that the tree has not yet visited. (d) Acting by selecting the node with the highest weight among the child nodes for decision-making. (e) Pruning of the tree to synchronize with the environment.}
   \label{fig:tree_details}
   \end{center}
   \vskip -0.15in
\end{figure*}

This section describes how to construct the TAT and plan with it, incorporating details from Figure~\ref{fig:tree_details}.

\textbf{Initialization.}~~
Initially, we create the foundation for our TAT, symbolized as $\varUpsilon$. This step involves establishing an empty root node $e_{0,0}$, where $X(e_{0,0})$ starts empty, indicating no data is yet collected. $V(e_{0,0})$ is also empty showing no weights assigned. $\bm{x}(e_{0,0})$ is set to $\bm{0}$, representing the initial node state.
Next, TAT evolves synchronized with environmental steps. If new trajectories emerge, they are integrated into the existing $\varUpsilon$. Integration of a trajectory $\bm{\tau} = \{\bm{x}_0, \dots, \bm{x}_T\}$ involves two key steps: (1) \textbf{Merging} similar branch to consolidate common patterns and (2) \textbf{Expanding} different branch to accommodate new information.

\textbf{Merging.}~~
This process aims to merge the trajectory elements $\bm{x}_t$ into the nodes $e_{i,t}$ with similar state characteristics, following a path beginning at the root node $e_{0,0}$. The initial state $\bm{x}_0$ is merged directly into $e_{0,0}$ since it has no impact on decision-making. At each subsequent time step $t$, we evaluate the child nodes ${e_{t,i}}$ of the last merged node at depth $t-1$, seeking a node whose state closely aligns with $\bm{x}_t$. For discrete states, this involves checking for exact matches, while for continuous states, cosine similarity, $\mathfrak{S}^{\rm cos}(\bm{x}_t, e_{t,i})=\frac{\bm{x}_t \cdot \bm{x}(e_{t,i})}{\|\bm{x}_t\|\cdot\|\bm{x}(e_{t,i})\|}$, is calculated to identify the most suitable node. A threshold $\alpha$, marginally less than one\footnote{The value of cosine similarity is [-1,1], and the more similar the two states are, the closer the cosine similarity is to 1.}, is employed to guarantee adequate similarity. The node with the highest similarity while satisfying $\mathfrak{S}^{\rm cos}>\alpha$ is selected for merging. Then, the visited node receives an additional weight and updates its statistics:
\begin{equation}
   \label{eq:update_statistics_1}
   X(e_{t,i}).\texttt{insert}(\bm{x}_t), V(e_{t,i}).\texttt{insert}(\lambda^t),
\end{equation}
\begin{equation}
   \label{eq:update_statistics_2}
   \bm{x}(e_{t,i}) = \frac{\sum_{\{\bm{x}_i,\lambda^i\}\in \left\{X(e_{t,i}),V(e_{t,i})\right\}}{\bm{x}_i\cdot\lambda^i}}{\sum_{\lambda^i\in V(e_{t,i})}\lambda^i},
\end{equation}
where Equation~(\ref{eq:update_statistics_1}) represents the inclusion of the new state $\bm{x}_t$ and its weight $\lambda^t$, thus enhancing the node's significance, with ${\lambda^t}$ denoting the weight assigned to each node by a decay factor $\lambda$ (where $\lambda \leq 1$), prioritizing recent states for predictive accuracy, as shown in Figure~\ref{fig:tree_details} (a). Equation~(\ref{eq:update_statistics_2}) updates the node state $\bm{x}(e_{t,i})$ to the weighted average of its states, ensuring dynamic adaptation to new data.

\textbf{Expanding.}~~
The merging procedure is terminated (presumably at step $L$) when there are no suitable nodes for transition, e.g., when $\mathfrak{S}^{\rm cos}(\bm{x}_t, e_{t,i})\leq\alpha, \forall i\in \{0,1,\dots\}$. At this point, the latter segment of the trajectory, previously unexplored by the tree, remains unmerged. As such, the entire trajectory $\bm{\tau}$ is naturally divided into two parts: the former $\bm{\tau}^{\rm former}=\{\bm{x}_0,\dots, \bm{x}_{L-1}\}$ and the latter $\bm{\tau}^{\rm latter}=\{\bm{x}_L,\dots, \bm{x}_T\}$. The former sub-trajectory $\bm{\tau}^{\rm former}$ is the part that can be traced within the tree. Conversely, the latter sub-trajectory $\bm{\tau}^{\rm latter}$ represents the part that the tree has not yet visited before. Consequently, we undertake an expansion for the tree with the latter sub-trajectory $\bm{\tau}^{\rm latter}$, as demonstrated in Figure~\ref{fig:tree_details} (c). We initialize a node $\big\{X(e_{t,i})=\{\bm{x}_t\},V(e_{t,i})=\{\lambda^t\}, \bm{x}(e_{t,i})=\bm{x}_t\big\}$ for each state $\bm{x}_t\in\bm{\tau}^{\rm latter}$, where $t=L,\dots,T$. Then, we attach the latter state's node as a child node of the previous state's node, sequentially expanding the tree.

\textbf{Acting.}~~
In the current state (at the root node $e_{0,0}$), the next state needs to be chosen for decision-making. TAT makes its decisions
based on the most impactful node, which has the highest weight among the child nodes $\{e_{1,i}\}_{i=0,1,\dots}$:
\begin{equation}
\label{eq:action_selection}
   e^{\ast}_1=\mathop{\arg\max}_{e\in\{e_{1,i}\}_{i=0,1,\dots}}Q(e),
\end{equation}
where $Q(e)=\sum\nolimits_{\lambda^i\in V(e)}\lambda^i$ is the accumulated weight of node $e$. We then take its represented state $\bm{x}^{\ast}_1=\bm{x}(e^{\ast}_1)$ as our target next state. To decide the action $\textbf{a}^{\ast}_0$ for the current environment state $\textbf{s}_0$, we adopt the original strategies from prior methods. If it is the state-action pair, like Diffuser~\cite{janner2022planning}, the action is directly derived from $\bm{x}^{\ast}_1=(\textbf{s}^{\ast}_1,\textbf{a}^{\ast}_0)$; If it is the state-centric form ($\bm{x}^{\ast}_1=\textbf{s}^{\ast}_1$), like Decision Diffuser~\cite{ajay2023is}, the action is derived via an inverse dynamics model $\textbf{a}^{\ast}_0 = f_{\phi}(\textbf{s}_0, \textbf{s}^{\ast}_{1})$.

\textbf{Pruning.}~~
After making the decision, the environment transitions to the next state, and the tree is pruned to reflect this transition. The subtree rooted at the selected node $e^{\ast}_1$ becomes the new tree:
$e_{0,0} \gets e^{\ast}_1$.
As such, the tree is dynamically updated over time with the progression of the decision-making process, ensuring that the model's predictions are in sync with the environment state transitions.

Pseudocode of closed-loop planning with TAT in a single episode is given in Algorithm~\ref{alg:tree_planning}, where lines 11-18 correspond to the merging, lines 20-23 correspond to the expanding, line 25 corresponds to the acting, and line 27 corresponds to the pruning. Open-loop planning can be easily implemented by sampling trajectories (lines 5-9) just once within the initial while loop. 

\begin{algorithm}[t]
   \caption{Planning with TAT}
   \label{alg:tree_planning}
   \begin{algorithmic}[1]
     \STATE {\bfseries Require:} Pretrained diffusion model $g_{\theta}$
     \STATE Initialize an empty tree $\varUpsilon \gets e_{0,0}$
     \FOR{each step of the environment}
     \STATE \small{\color{gray}{// Sampling plans via diffusion}}
     \STATE Observe current state $\textbf{s}$; Initialize plan $\bm{\tau}_K\sim \mathcal{N}(\bm{0},\textbf{I})$
     \FOR{$k=K,\dots,1$}
     \STATE Set $\textbf{s}$ as the first state of $\bm{\tau}_k$
     \STATE Perform one-step denoising: $\bm{\tau}_{k-1} \gets g_{\theta}(\bm{\tau}_k,k)$ 
     \ENDFOR
     \STATE \small{\color{gray}{// Merging}}
     \FOR{$t=1,\dots$}
     \STATE Compute cosine similarities $\{\mathfrak{S}^{\rm cos}(\bm{x}_t, e_{t,i})\}_{i=1=0,1,\dots}$
     \IF{$\forall i\in \{0,1,\dots\}, \mathfrak{S}^{\rm cos}(\bm{x}_t, e_{t,i})\leq\alpha$}
     \STATE Break the loop (presumably at $t=L$)
     \ENDIF
     \STATE Select the node with the highest $\mathfrak{S}^{\rm cos}$
     \STATE Merge $\bm{x}_t$ into the selected node by Equation~\eqref{eq:update_statistics_1}, \eqref{eq:update_statistics_2}
     \ENDFOR
     \STATE \small{\color{gray}{// Expanding}}
     \FOR{$t=L,\dots, T$}
     \STATE Initialize a node with state $\bm{x}_t$
     \STATE Expand $\varUpsilon$ by attaching this node as a new leaf node
     \ENDFOR
     \STATE \small{\color{gray}{// Acting}}
     \STATE Execute action $\textbf{a}_0$ drawn from $e^{\ast}_1$ in Equation~\eqref{eq:action_selection}
     \STATE \small{\color{gray}{// Pruning}}
     \STATE Transition to the next state and prune the tree $e_{0,0} \gets e^{\ast}_1$
     \ENDFOR
   \end{algorithmic}
 \end{algorithm}

\subsection{Theoretical Analysis}
\begin{figure}[t]
   \vskip 0.1in
   \begin{center}
   \centerline{\includegraphics[width=0.85\columnwidth]{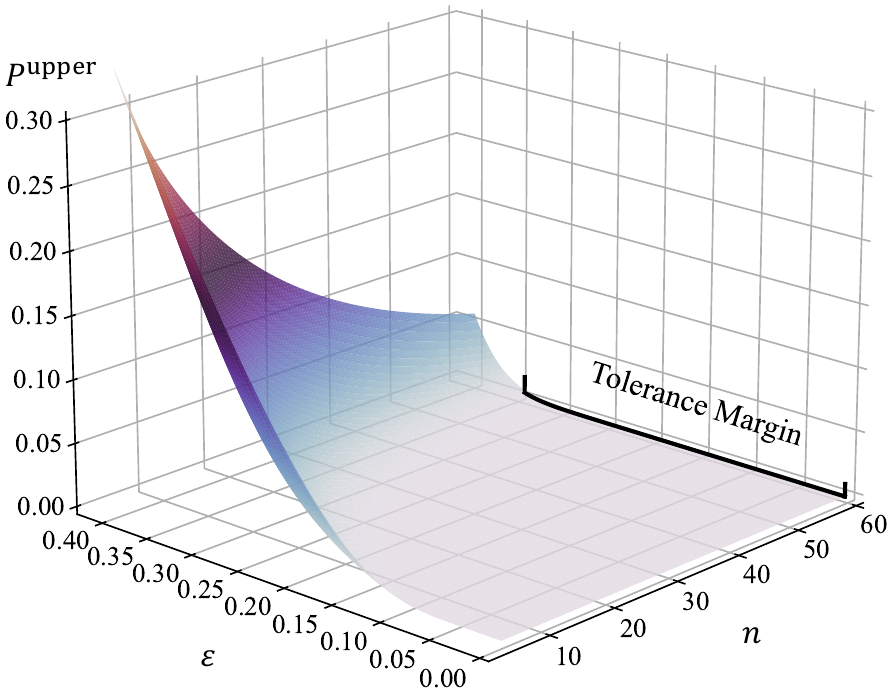}}
   \caption{The dynamic trend of the upper bound $P^{\text{upper}}$ of $P^{\text{artifact}}(n; \varUpsilon)$ with respect to $\varepsilon$ and $n$. As $n$ increases, it monotonically decreases and becomes less sensitive to the variation of $\varepsilon$.}
   \label{fig:theoretical_p_artifact}
   \end{center}
   \vskip -0.15in
\end{figure}

In this section, we provide a theoretical analysis of TAT's effectiveness in the presence of artifacts. All the proofs pertinent to this section are included in Appendix~\ref{sec:apendix_proofs}. We first give some key definitions.
\begin{definition}[Artifact Probability]
We let $\varepsilon$ denote the probability of each $x_t$ being an artifact.
\end{definition}
\begin{definition}[Number of Trajectories]
We let $n$ denote the total number of trajectories involved across all candidate child nodes $\{e_{1,i}\}_{i=0,1,…}$ in Equation~(\ref{eq:action_selection}), i.e., $n=\sum_{i=0,1,\dots}|X(e_{1,i})|$. Hence, $n$ is a positive integer greater than 1, i.e., $n\in \mathbb{Z}^{+}_{>1}$.
\end{definition}
We assume independence in the generation of artifacts across different trajectories, which generally holds as trajectories are derived from i.i.d. Gaussian noise~\cite{sohl2015deep}. We also assume that the diffusion planner has practical significance, leaning towards producing reliable states rather than unreliable ones, i.e., $\varepsilon<0.5$. We proceed to introduce a proposition that involves the bounded probability of TAT choosing an artifact for decision-making.
\begin{proposition}\label{prop:probability_bound}
   In TAT's planning, each action is a product of aggregated information from $n$ trajectories. Consider the case of $\lambda=1$, the probability of TAT choosing the artifact can be bounded by
   \begin{equation}
      P^{\text{\rm artifact}}(n; \varUpsilon) < \frac{1}{2}\left[1-{\rm erf}\left(\frac{n/2  - n\varepsilon}{\sqrt{2 n\varepsilon(1-\varepsilon)}}\right)\right], \nonumber
   \end{equation}
   where ${\rm erf}(\cdot)$ denotes the error function.
\end{proposition}
Proposition~\ref{prop:probability_bound} implies that as the number of trajectories $n$ increases, the term inside the error function grows, which in turn leads to a monotonically decreasing upper bound of $P^{\text{artifact}}(n; \varUpsilon)$. Hence, involving more trajectories in the system becomes a feasible strategy to enhance the performance. Besides, we can further derive the following corollary.
\begin{corollary}\label{cor:probability_properties}
For $\forall n\in \mathbb{Z}^{+}_{>1}$, the probability of TAT selecting an artifact satisfies $P^{\text{\rm artifact}}(n; \varUpsilon) < \varepsilon$ and $\lim_{n \to \infty} P^{\text{\rm artifact}}(n; \varUpsilon) = 0$.
\end{corollary}
Corollary~\ref{cor:probability_properties} highlights two important properties of $P^{\text{artifact}}(n; \varUpsilon)$. Firstly, TAT's probability of selecting an artifact is always lower than that of an individual trajectory, ensuring performance improvement. Secondly, with a sufficiently large number of trajectories, the likelihood approaches zero. In Figure~\ref{fig:theoretical_p_artifact}, we provide a visual illustration of the dynamic trend of the upper bound to support our theoretical claims. Moreover, we observe that when $n$ is large (e.g., $n=60$), the upper bound is not sensitive to variations in $\varepsilon$ within a wide interval (e.g., $\varepsilon \in [0.00,0.35]$). This observation indicates that TAT manifests a tolerance margin for the generation of artifacts, which allows diffusion planners to achieve faster yet reliable planning by reducing sampling steps.

\section{Experiments}
In this section, we present the empirical evaluations of the proposed TAT across a range of decision-making tasks in offline control settings. 
As our TAT serves as plug-and-play and training-free enhancement, we use the open-source pre-trained models of existing diffusion planners, deploying TAT directly. In cases where pre-trained models are not publicly available, we retrain them using the original hyperparameters. For more detailed information, please refer to Appendix~\ref{sec:appendix_experimental_details}. Our experiments are designed to demonstrate \textbf{(1)} the ability of TAT to filter out potential artifacts and improve the reliability of the planner; \textbf{(2)} its ability to boost the performance of diffusion planners as a plug-and-play enhancement; \textbf{(3)} its ability to achieve faster yet reliable planning benefiting from its tolerance for artifact generation. Source code is available at \url{https://github.com/langfengQ/tree-diffusion-planner}.

\subsection{Risk Resistance}
In this part, we deploy our TAT ($\varUpsilon$) on the pre-trained Diffuser~\cite{janner2022planning} and denote it as $\text{Diffuser}^{\varUpsilon}$. We evaluate TAT on the Maze2D environments~\cite{fu2020d4rl} to show its effectiveness in minimizing the artifacts' risks in the original Diffuser. Maze2D presents a navigation task, challenging the agent to plan a path to reach a designated goal location, where only a reward of 1 is granted. The sampling procedure of Diffuser in Maze2D tasks is conditioned on a start and goal location without return guidance.

\begin{figure*}[t]
   \vskip 0.1in
   \begin{center}
   \centerline{\includegraphics[width=1.0\textwidth]{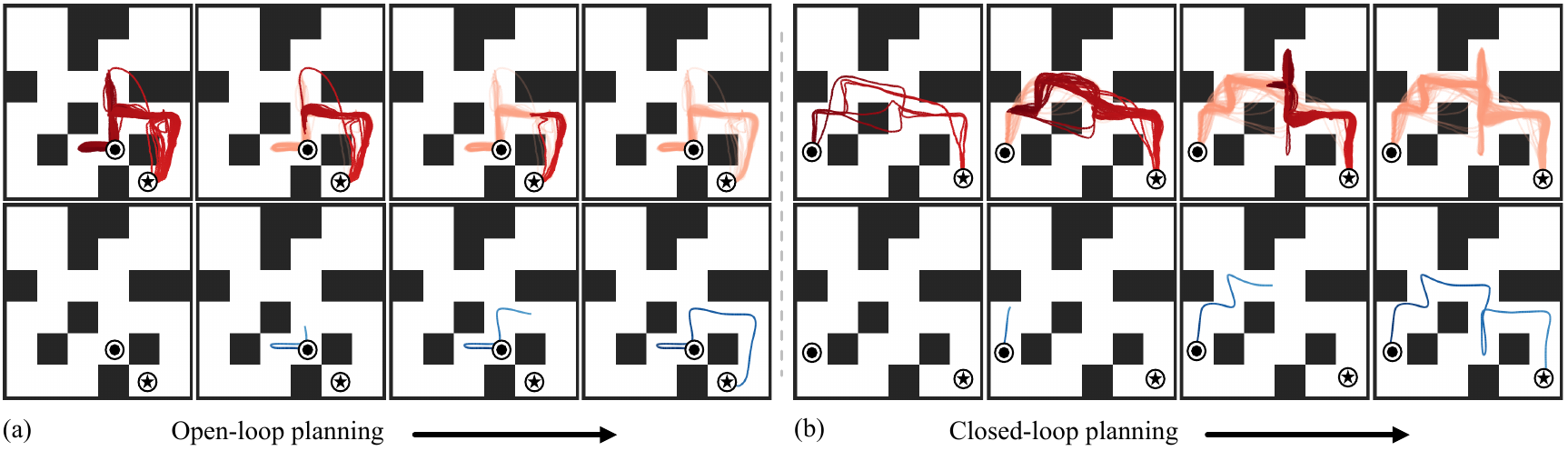}}
   \caption{The planning process of $\text{Diffuser}^{\varUpsilon}$ in open-loop (a) and closed-loop (b) manners. \protect{\raisebox{-.05cm}{\includegraphics[height=.3cm]{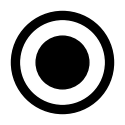}}} denotes the starting position and \protect{\raisebox{-.05cm}{\includegraphics[height=.3cm]{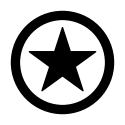}}} denotes the goal position. The red lines on the top row are the plans of the baseline Diffuser. It is evident that Diffuser generates artifacts at the beginning (a) or introduces new artifacts in subsequent steps (b). $\text{Diffuser}^{\varUpsilon}$ can dynamically filter out and prune these inefficient branches indicated by the light transparent segments. The blue lines on the below row are the results of $\text{Diffuser}^{\varUpsilon}$.}
   \label{fig:risk_diversification}
   \end{center}
   \vskip -0.2in
\end{figure*}

As shown in Figure~\ref{fig:risk_diversification}, Diffuser inevitably produces some infeasible transitions in both open-loop and closed-loop planning, such as instances of wall-crossing, with 5 out of 64 trajectories in Figure~\ref{fig:risk_diversification} (a). Therefore, it is risky and unreliable to plan with an individual trajectory in prior methods. In contrast, $\text{Diffuser}^{\varUpsilon}$ aggregates multiple trajectories, dynamically expands and prunes the tree, and makes decisions based on the most impactful nodes. In Figure~\ref{fig:risk_diversification}, $\text{Diffuser}^{\varUpsilon}$ filters out all impractical transitions and exhibits a strong ability to resist the stochastic risk from artifacts. As indicated by the blue paths, $\text{Diffuser}^{\varUpsilon}$ offers a more reliable and artifact-free path towards the goal location.

\begin{table}[t]
   \caption{The performance of $\text{Diffuser}^{\varUpsilon}$, Diffuser, and varieties of prior approaches in the Maze2D environment. We report the mean and the standard error of $\text{Diffuser}^{\varUpsilon}$ over 1000 planning seeds.}
   \label{tab:maze2d_result}
   \vskip 0.1in
   \begin{center}
   \begin{small}
   % \begin{sc}
      \begin{tabular}{@{\,\,}l@{\,\,}c@{\,\,\,\,}c@{\,\,\,\,}c@{\,\,\,\,}c@{\,\,\,\,}l@{\,}}
         \toprule
         \textbf{Environment} & \textbf{MPPI} & \textbf{CQL} & \textbf{IQL} & \textbf{Diffuser} & $\textbf{Diffuser}^{\varUpsilon}$ \\
         \midrule
         Maze2d \ \ U-Maze   & 33.2            & \phantom{0} 5.7 & 47.4  & \textbf{113.9} & \textbf{114.5}{\tiny$\pm$1.2}   \\
         Maze2d \ \ Medium   & 10.2            & \phantom{0} 5.0 & 34.9  & 121.5          & \textbf{130.7}{\tiny$\pm$0.6}   \\
         Maze2d \ \ Large    & \phantom{0}5.1  & 12.5            & 58.6  & 123.0          & \textbf{133.4}{\tiny$\pm$1.7} \\
         \midrule
         \textbf{Single-task Average} & 16.2   & \phantom{0}7.7  & 47.0  & 119.5 & \textbf{126.2}\\
         \midrule
         Multi2d \ \ U-Maze   & 41.2           & -               & 24.8  & \textbf{128.9} & \textbf{129.4}{\tiny$\pm$0.6}   \\
         Multi2d \ \ Medium   & 15.4           & -               & 12.1  & 127.2           & \textbf{135.4}{\tiny$\pm$0.9}   \\
         Multi2d \ \ Large    &\phantom{0}8.0  & -               & 13.9  & 132.1          &  \textbf{143.8}{\tiny$\pm$1.5} \\
         \midrule
         \textbf{Multi-task Average} &21.5     & -               & 16.9  & 129.4    & \textbf{136.2}\\
         \bottomrule
   \end{tabular}
   % \end{sc}
   \end{small}
   \end{center}
   \vskip -0.15in
\end{table}

\subsection{Performance Boosting}
We evaluate TAT's ability to boost the performance of existing diffusion planners in three different offline RL tasks.

\textbf{Maze2D.}~~
We first evaluate our method on sparse-reward, long-horizon tasks in Maze2D~\cite{fu2020d4rl}. As we describe above, it is a particularly demanding navigation task as the agent often needs to take hundreds of steps without receiving any rewards until reaching the final goal. Successfully navigating the maze requires the agent to adeptly handle sparse rewards and exhibit proficiency in long-horizon planning. We also assess multi-task flexibility on Multi2D, a variant of Maze2D that introduces goal randomization for each episode~\cite{janner2022planning}. We compare our TAT-reinforced $\text{Diffuser}^{\varUpsilon}$ with original Diffuser~\cite{janner2022planning} and other baselines in Table~\ref{tab:maze2d_result}.

Notably, model-based trajectory optimizer MPPI~\cite{williams2015model} and model-free offline algorithms CQL~\cite{kumar2020conservative} and IQL~\cite{kostrikov2022offline} all struggle to reach the goal. The state-of-the-art model-free IQL exhibits a significant performance drop in multi-task navigation due to the great difficulties in performing credit assignments. In contrast, Diffuser performs much better in both single-task and multi-task scenarios, but its performance is affected by artifacts shown in Figure~\ref{fig:risk_diversification}. Our TAT effectively addresses this risk and boosts the performance of the Diffuser, particularly in larger and more complex mazes, where the likelihood of generating artifacts increases. We can observe an improvement of around 10 scores in such cases.

\begin{table}[t]
   \caption{The performance of $\text{Diffuser}^{\varUpsilon}$, Diffuser and various prior approaches on block stacking tasks. We report the mean and the standard error of $\text{Diffuser}^{\varUpsilon}$ and $\text{DD}^{\varUpsilon}$ over 100 planning seeds.}
   \label{tab:block_stacking_result}
   \vskip 0.1in
   \begin{center}
   \begin{small}
   % \begin{sc}
      \begin{tabular}{@{\,}l@{\,\,}c@{\,\,\,}c@{\,\,\,}c@{\,\,}l@{\,\,}c@{\,\,\,\,}l@{\,}}
         \toprule
         \textbf{Environment} & \textbf{BCQ} & \textbf{CQL} & \textbf{Diffuser} & $\textbf{Diffuser}^{\varUpsilon}$ & \textbf{DD} & $\textbf{DD}^{\varUpsilon}$\\
         \midrule
         Unconditional   & 0.0 & 24.4            & 54.3   & \textbf{60.7}{\tiny$\pm$2.3} & 56.0   & \textbf{63.7}{\tiny$\pm$2.6} \\
         Conditional     & 0.0 & \phantom{0}0.0  & 48.7   & 56.7{\tiny$\pm$3.5} & 54.0   & \textbf{62.3}{\tiny$\pm$3.5} \\
         Rearrangement            & 0.0 & \phantom{0}0.0  & 51.3   & 63.3{\tiny$\pm$3.0} & 59.8   & \textbf{69.2}{\tiny$\pm$3.1}\\
         \midrule
         \textbf{Average}         & 0.0 & \phantom{0}8.1  & 51.4   & 60.2 & 56.6   & \textbf{65.1} \\
         \bottomrule
   \end{tabular}
   % \end{sc}
   \end{small}
   \end{center}
   \vskip -0.15in
\end{table}

\begin{table*}[t]
   \caption{The performance of various diffusion planners with/without TAT, and varieties of prior approaches on MuJoCo locomotion tasks. We report the normalized average returns and the standard errors of $\text{Diffuser}^{\varUpsilon}$ and $\text{RGG}^{\varUpsilon}$ over 50 planning seeds. We highlight scores that fall within 5 percent of the maximum per task ($\geq0.95\cdot\text{max}$).}
   \label{tab:locomotion_result}
   \vskip 0.1in
   \begin{center}
   \begin{small}
   % \begin{sc}
      \begin{tabular}{@{\,\,}l@{\,\,\,}l@{\,\,\,}c@{\,\,\,\,}c@{\,\,\,\,}c@{\,\,\,\,}c@{\,\,\,\,}c@{\,\,\,}c@{\,\,\,}c@{\,\,\,}c@{\,\,\,}c@{\,\,\,}c@{\,\,\,}c@{\,\,\,\,}c@{\,\,}}
         \toprule
         \textbf{Dataset} & \textbf{Environment} & \textbf{BC} & \textbf{CQL} & \textbf{IQL} & \textbf{DT} & \textbf{TT} & \textbf{MOPO} & \textbf{MOReL} & \textbf{MBOP} & \textbf{Diffuser} & $\textbf{Diffuser}^{\varUpsilon}$ & \textbf{RGG} & $\textbf{RGG}^{\varUpsilon}$ \\
         \midrule
         Med-Expert & HalfCheetah & \phantom{0}55.2 & \phantom{0}91.6 & \phantom{0}86.7  & \phantom{0}86.8  & \phantom{0}95.0  & \phantom{0}63.3  & \phantom{0}53.3  & \textbf{105.9} & \phantom{0}88.9  & \phantom{0}92.5{\tiny$\pm$0.8}   & \phantom{0}90.8  & \phantom{0}95.6{\tiny$\pm$1.1}\\
         Med-Expert & Hopper      & \phantom{0}52.5 & 105.4           & \phantom{0}91.5  & \textbf{107.6} & \textbf{110.0}  & \phantom{0}23.7  & \textbf{108.7} & \phantom{0}55.1  & 103.3 & \textbf{109.4}{\tiny$\pm$2.1}   & \textbf{109.6} & \textbf{111.2}{\tiny$\pm$1.8}\\
         Med-Expert & Walker2d    & \textbf{107.5}  & \textbf{108.8}  & \textbf{109.6}   & \textbf{108.1} & 101.9 & \phantom{0}44.6  & \phantom{0}95.6  & \phantom{0}70.2  & \textbf{106.9} & \textbf{108.8}{\tiny$\pm$0.3}   & \textbf{107.8} &  \textbf{109.3}{\tiny$\pm$0.2}\\
         \midrule
         Medium & HalfCheetah     & \phantom{0}42.6 & \phantom{0}44.0 & \phantom{0}\textbf{47.4}  & \phantom{0}42.6  & \phantom{0}\textbf{46.9}  & \phantom{0}42.3  & \phantom{0}42.1  & \phantom{0}44.6  & \phantom{0}42.8  & \phantom{0}44.3{\tiny$\pm$0.3}   & \phantom{0}44.0  & \phantom{0}\textbf{45.7}{\tiny$\pm$0.4}\\
         Medium & Hopper          & \phantom{0}52.9 & \phantom{0}58.5 & \phantom{0}66.3  & \phantom{0}67.6  & \phantom{0}61.1  & \phantom{0}28.0  & \phantom{0}\textbf{95.4}  & \phantom{0}48.8  & \phantom{0}74.3  & \phantom{0}82.6{\tiny$\pm$1.9} & \phantom{0}82.5  & \phantom{0}86.0{\tiny$\pm$2.3}\\
         Medium & Walker2d        & \phantom{0}75.3 & \phantom{0}72.5 & \phantom{0}78.3 & \phantom{0}74.0  & \phantom{0}\textbf{79.0}  & \phantom{0}17.8  & \phantom{0}77.8  & \phantom{0}41.0  & \phantom{0}\textbf{79.6}  & \phantom{0}\textbf{81.0}{\tiny$\pm$0.4}  & \phantom{0}\textbf{81.7}  & \phantom{0}\textbf{82.7}{\tiny$\pm$0.8}\\
         \midrule
         Med-Replay & HalfCheetah & \phantom{0}36.6 & \phantom{0}45.5 & \phantom{0}44.2  & \phantom{0}36.6  & \phantom{0}41.9  & \phantom{0}\textbf{53.1}  & \phantom{0}40.2  & \phantom{0}42.3  & \phantom{0}37.7  & \phantom{0}39.2{\tiny$\pm$1.5}   & \phantom{0}41.0  & \phantom{0}43.6{\tiny$\pm$1.6}\\
         Med-Replay & Hopper      & \phantom{0}18.1 & \phantom{0}\textbf{95.0} & \phantom{0}\textbf{94.7} & \phantom{0}82.7  & \phantom{0}91.5  & \phantom{0}67.5  & \phantom{0}93.6  & \phantom{0}12.4  & \phantom{0}93.6  & \phantom{0}\textbf{95.3}{\tiny$\pm$0.3}    & \phantom{0}\textbf{95.2}  & \phantom{0}\textbf{99.5}{\tiny$\pm$0.5}\\
         Med-Replay & Walker2d    & \phantom{0}26.0 & \phantom{0}77.2 & \phantom{0}73.9  & \phantom{0}66.6  & \phantom{0}\textbf{82.6}  & \phantom{0}39.0  & \phantom{0}49.8  &  \phantom{0}\phantom{0}9.7  & \phantom{0}70.6  & \phantom{0}78.2{\tiny$\pm$2.2}    & \phantom{0}78.3  & \phantom{0}\textbf{82.3}{\tiny$\pm$1.9}\\
         \midrule
         \textbf{Average} &       & \phantom{0}51.9 & \phantom{0}77.6 & \phantom{0}77.0  & \phantom{0}74.7  & \phantom{0}78.9  & \phantom{0}42.1  & \phantom{0}72.9  & \phantom{0}47.8  & \phantom{0}77.5  & \phantom{0}\textbf{81.3}\phantom{\tiny$\pm$0.0}  & \phantom{0}\textbf{81.2} & \phantom{0}\textbf{84.0}\phantom{\tiny$\pm$0.0} \\
         \bottomrule
   \end{tabular}
   % \end{sc}
   \end{small}
   \end{center}
   \vskip -0.15in
\end{table*}

\textbf{Kuka block stacking.}~~
The Kuka block stacking suite~\cite{janner2022planning} is designed for evaluating algorithms' test-time flexibility. It involves the manipulation of a Kuka robotic arm~\cite{schreiber2010fast} to accomplish three tasks: \emph{unconditional stacking} to build the tallest possible block tower, \emph{conditional stacking} to build a block tower in a specified order, and \emph{rearrangement} to match the locations of reference blocks in a new arrangement. 
We deploy TAT on Diffuser~\cite{janner2022planning} and Decision Diffuser (DD)~\cite{ajay2023is} and the sampling is guided by specific task conditions. The results are given in Table~\ref{tab:block_stacking_result}.

Compared to model-free offline RL algorithms BCQ~\cite{fujimoto2019off} and CQL~\cite{kumar2020conservative}, diffusion planners exhibit an ability to adapt to different block properties and stacking criteria at test time. Notably, $\text{Diffuser}^{\varUpsilon}$ and $\text{DD}^{\varUpsilon}$ demonstrates a clear improvement margin ($>15\%$ overall) over Diffuser and DD, with more significant enhancements in tasks that involve stacking constraints.

\textbf{MuJoCo locomotion.}~~
At last, we evaluate our method on MuJoCo tasks using D4RL offline locomotion suite~\cite{fu2020d4rl}, which involves policy mixtures to assess algorithms' capability on heterogeneous data with varying quality. We choose pre-trained Diffuser~\cite{janner2022planning} and Restoration Gap Guidance (RGG)~\cite{lee2023refining} as baseline diffusion planners, and equip them with TAT to form $\text{Diffuser}^{\varUpsilon}$ and $\text{RGG}^{\varUpsilon}$, respectively. In these tasks, diffusion planners typically adopt closed-loop planning and are guided by high-return conditions. In addition, we also compared our method against various other baselines known for their strong performance in each domain of tasks, including imitation-based method behavior cloning (BC); model-free offline RL algorithms CQL~\cite{kumar2020conservative} and IQL~\cite{kostrikov2022offline}; sequence modeling method Decision Transformer (DT)~\cite{chen2021decision}; model-based RL methods Trajectory Transformer (TT)~\cite{janner2021offline}, MOPO~\cite{yu2020mopo}, MOReL~\cite{kidambi2020morel}, MBOP~\cite{argenson2021modelbased}.
The results are presented in Table~\ref{tab:locomotion_result}. 

It can be seen that TAT-reinforced planners demonstrate better performance compared to model-based approaches like MOReL and MBOP, as well as the sequence modeling approach DT. In addition, while the original Diffuser falls short in performance when compared to offline RL algorithms like CQL and TT, our TAT enables it to surpass these strong algorithms. Notably, TAT-reinforced planners ($\text{Diffuser}^{\varUpsilon}$ and $\text{RGG}^{\varUpsilon}$) consistently outperform their respective baseline planners in all tasks, without instances of performance drop. In some tasks, the performance gains can exceed $10\%$.

\subsection{Faster Planning}
\begin{table}[t]
   \caption{The planning time and performance of $\text{Diffuser}^{\varUpsilon}$ on Hopper Medium task with fewer sampling steps. The planning time of each action is averaged over 200 actions' generation and the score is averaged over 50 planning seeds. Originally set at 20 steps, the default sampling for Diffuser yields baseline results of 1.68s (time) and 74.3 (score). The symbols \ding{51} and \ding{55} are used to indicate whether $\text{Diffuser}^{\varUpsilon}$ performs better or worse than corresponding baseline results, respectively.}
   \label{tab:warm_start}
   \vskip 0.1in
   \begin{center}
   \begin{small}
   % \begin{sc}
   \begin{tabular}{@{\,\,}c|cc|cc@{\,\,}}
         \toprule
         \multirow{2}{*}{\textbf{Step}} & \multicolumn{2}{c|}{\textbf{Warm-Start Sampling}}  & \multicolumn{2}{c}{\textbf{$\text{Warm-Start}^{\varUpsilon}$ Sampling}}  \\
                     & \rule{0pt}{2.6ex}Time  &Score & Time  & Score \\
         \midrule
         16           & 1.43s~~\ding{51}   &  79.6{\tiny$\pm$2.4}~~\ding{51}   & 1.44s~~\ding{51}    & 81.0{\tiny$\pm$2.4}~~\ding{51}  \\
         12           & 1.08s~~\ding{51}   &  79.9{\tiny$\pm$2.2}~~\ding{51}   & 1.09s~~\ding{51}   & 82.7{\tiny$\pm$1.9}~~\ding{51}       \\
         10            & 0.91s~~\ding{51}   &  79.4{\tiny$\pm$2.2}~~\ding{51}   & 0.92s~~\ding{51} & 82.6{\tiny$\pm$1.9}~~\ding{51}  \\
         8            & 0.75s~~\ding{51}   &  77.2{\tiny$\pm$2.3}~~\ding{51}   & 0.76s~~\ding{51} & 82.4{\tiny$\pm$1.7}~~\ding{51}  \\
         6            & 0.53s~~\ding{51}   &  71.2{\tiny$\pm$2.2}~~\ding{55}   &0.55s~~\ding{51}     & 79.6{\tiny$\pm$2.0}~~\ding{51} \\
         4            & 0.41s~~\ding{51}   &  58.9{\tiny$\pm$1.2}~~\ding{55}   & 0.43s~~\ding{51}    & 64.8{\tiny $\pm$1.4}~~\ding{55} \\
         % 2            & \textbf{0.25}   &  43.9{\tiny$\pm$2.4}            & \textbf{0.27}    & 50.1{\tiny$\pm$1.4} \\
         \bottomrule
   \end{tabular}
   % \end{sc}
   \end{small}
   \end{center}
   \vskip -0.15in
\end{table}
In Proposition~\ref{prop:probability_bound} and Figure~\ref{fig:theoretical_p_artifact}, we demonstrated that TAT exhibits a tolerance margin for the generation of artifacts. Moreover, Figure~\ref{fig:risk_diversification} also illustrates TAT's ability to manage a growing proportion of infeasible trajectories, effectively filtering them out. This indicates that we can strategically sacrifice the sample quality by reducing denoising steps for faster planning. Such an aspect holds particular significance for diffusion planners, which are commonly subject to extensive time requirements from iterative denoising procedures.

To validate this, we trade sample quality for reduced denoising steps by warm-start planning~\cite{janner2022planning}. This strategy leverages previously generated plans to create partially noised trajectories, which are then used as starting points for subsequent planning iterations to reduce denoising steps. We present the results of $\text{Diffuser}^{\varUpsilon}$ in Table~\ref{tab:warm_start}. We observed that $\text{Diffuser}^{\varUpsilon}$ maintains consistent performance for sampling steps exceeding 8. Within this range, we can accelerate the planning of $\text{Diffuser}^{\varUpsilon}$ while ensuring it outperforms the original Diffuser. Moreover, TAT can optimize this process by rollouting branches with the highest weight for warm-start planning, denoted as $\text{warm-start}^{\varUpsilon}$ sampling. We find that this technique not only enhances the robustness of $\text{Diffuser}^{\varUpsilon}$ to variations in sampling steps but also marginally improves performance. We present the results in Table~\ref{tab:warm_start}. Remarkably, $\text{Diffuser}^{\varUpsilon}$ achieves better results than vanilla warm-start planning. It consistently outperforms Diffuser when the sampling step is reduced to just 6, significantly decreasing planning latency to 0.55s (more than a threefold increase in speed). In Appendix~\ref{sec:planning_time}, we also present the time and memory budget of TAT, which we find to be quite modest.

% ~~\ding{51}
% ~~\ding{55}

\section{Conclusion}
In this work, we addressed the challenge of stochastic risk from infeasible trajectories in diffusion planners. We proposed a flexible and easy-to-deploy solution, TAT, that aggregates the collective wisdom of past and immediate trajectories. TAT presents a robust and practical structure to mitigate the impact of potential artifacts, and effectively exclude them from decision-making. Theoretical analysis and empirical results both provide support for the effectiveness of TAT. Remarkably, TAT consistently outperforms its baseline counterparts in all tasks and enables a remarkable increase in planning speed. Although our work focuses on the uncertainty risk of diffusion in decision-making domains, the concept of TAT could readily be extended to other areas, where similar stochasticity of generative models poses a challenge. Additionally, more sophisticated weight allocation and tree structures, such as incorporating spatial consideration, could be explored for further improvements. We leave it for future work.

\section*{Acknowledgements}
This research is supported by Natural Science Foundation of China (No. 61925603, U1909202), STI 2030 Major Projects (2021ZD0200400), and the Key Research and Development Program of Zhejiang Province in China (2020C03004). This research is also supported by the National Research Foundation, Singapore under its Industry Alignment Fund-Pre-positioning (IAF-PP) Funding Initiative. Any opinions, findings and conclusions or recommendations expressed in this material are those of the author(s) and do not reflect the views of National Research Foundation, Singapore.

\section*{Impact Statement}
This paper presents work whose goal is to advance the field of planning and decision-making. There are many potential societal consequences of our work, none of which we feel must be specifically highlighted here.

\bibliography{example_paper}
\bibliographystyle{icml2024}

%%%%%%%%%%%%%%%%%%%%%%%%%%%%%%%%%%%%%%%%%%%%%%%%%%%%%%%%%%%%%%%%%%%%%%%%%%%%%%%
%%%%%%%%%%%%%%%%%%%%%%%%%%%%%%%%%%%%%%%%%%%%%%%%%%%%%%%%%%%%%%%%%%%%%%%%%%%%%%%
% APPENDIX
%%%%%%%%%%%%%%%%%%%%%%%%%%%%%%%%%%%%%%%%%%%%%%%%%%%%%%%%%%%%%%%%%%%%%%%%%%%%%%%
%%%%%%%%%%%%%%%%%%%%%%%%%%%%%%%%%%%%%%%%%%%%%%%%%%%%%%%%%%%%%%%%%%%%%%%%%%%%%%%
\newpage
\appendix
\numberwithin{equation}{section}
\numberwithin{figure}{section}
\numberwithin{table}{section}
\numberwithin{algorithm}{section}
\onecolumn

\section{Guided Diffusion}\label{sec:appendix_guided_diffusion}
To sample trajectories that satisfy certain constraints, it is necessary to formulate the problem as a conditional distribution $p_{\theta}(\bm{\tau}|y(\bm{\tau}))$ using guided diffusion, where $y(\bm{\tau})$ is some specific condition on trajectory sample $\bm{\tau}$, e.g., the high-return $\mathcal{J}(\bm{\tau})$. There are two common choices to do so: classifier guidance~\cite{dhariwal2021diffusion} and classifier-free guidance~\cite{ho2021classifierfree}. The former combines the score estimate of a diffusion model with the gradient of $y(\bm{\tau})$ and thereby requires training a separate model $y_{\psi}$ to predict the scores of trajectory samples. The latter replaces the dedicated separate model with a diffusion model trained by randomly dropping the condition during training. 
The details of classifier guidance and classifier-free guidance for diffusion planners are as follows:

\subsection{Classifier Guidance}
Classifier guidance~\cite{dhariwal2021diffusion} modifies the denoising process by combining the score estimate of a diffusion model with the gradients of guidance functions. Therefore, it requires training auxiliary guidance functions $y_{\psi}$, which are trained separately. For instance, considering cumulative rewards $y_{\psi}=\mathcal{J}_{\psi}$, classifier guidance modifies the denoising process as follows:
\begin{equation}
p_{\theta}(\bm{\tau}_{k-1}|\bm{\tau}_k,y(\bm{\tau}))=\mathcal{N}(\bm{\tau}_{k-1};\mu_{\theta}(\bm{\tau}_{k},k)+\alpha\Sigma g,\Sigma),
\end{equation}
where $g=\nabla \mathcal{J}(\bm{\tau})|_{\bm{\tau}=\mu_{\theta}}$ and $\alpha$ is a guidance coefficient that controls the strength of the guidance. In this case, an additional model $\mathcal{J}_{\psi}$ is needed to be trained to predict the cumulative rewards of trajectory samples.

\subsection{Classifier-Free Guidance}
Classifier-free guidance~\cite{ho2021classifierfree} eliminates the need for separate guidance functions, with the diffusion model trained by randomly dropping the condition during training. Classifier-free guidance modifies the denoising process as follows:
\begin{align}
p_{\theta}(\bm{\tau}_{k-1}|\bm{\tau}_k,y(\bm{\tau}))=\mathcal{N}\left(\bm{\tau}_{k-1};\mu_{\theta}(\bm{\tau}_{k},k)+\omega(\mu_{\theta}(\bm{\tau}_{k},k,y(\bm{\tau}))-\mu_{\theta}(\bm{\tau}_{k},k)),\Sigma\right),
\end{align}
where $\omega$ is referred to as the guidance scale. $\mu_{\theta}(\bm{\tau}_{k},k,y(\bm{\tau}))$ is the mean of the conditional diffusion model, and $\mu_{\theta}(\bm{\tau}_{k},k)$ is the mean of the unconditional diffusion model. This method can be further simplified by applying it at the noise level:
\begin{align}
   \hat{\epsilon}_{\theta}(\bm{\tau}_k,k,y(\bm{\tau})) = \epsilon_{\theta}(\bm{\tau}_k,k)+\omega\left(\epsilon_{\theta}(\bm{\tau}_k,k,y(\bm{\tau}))-\epsilon_{\theta}(\bm{\tau}_k,k)\right),
\end{align}
where $\hat{\epsilon}_{\theta}(\bm{\tau}_k,k)$ is the perturbed noise, which will be used to later generate samples. $\epsilon_{\theta}(\bm{\tau}_k,k)$ and $\epsilon_{\theta}(\bm{\tau}_k,k,y(\bm{\tau}))$ are the unconditional and conditional noise models respectively.
\newpage

\section{Details of Diffusion Planners}\label{sec:appendix_diffuser_details}
In this section, we detail the planning process of diffusion planners. Generally, diffusion planning can be conducted in two ways: closed-loop planning and open-loop planning.

\paragraph{Closed-loop planning.} The planner generates a new plan $\bm{\tau}_0$ with a horizon of $T$ at each step of the environment. The plan is only used for decision-making at the current step. The planner can adjust the generated trajectory at each step based on environmental feedback to better adapt to the dynamic changes of the environment.
As an illustrative example, we present the workflow of closed-loop planning implemented in Diffuser~\cite{janner2022planning} in Algorithm~\ref{alg:diffuser_close_planning}.
\begin{algorithm}[h]
   \caption{Closed-loop planning with Diffuser}
   \label{alg:diffuser_close_planning}
   \begin{algorithmic}[1]
    \STATE \textbf{Input:} Noise model $\epsilon_{\theta}$, guide $\mathcal{J}$
    \FOR{each step of the environment}
        \STATE Observe current state $\textbf{s}$
        \STATE Initialize plan $\bm{\tau}_K\sim \mathcal{N}(\bm{0},\textbf{I})$
        \FOR{$k = K \dots 1$}
            \STATE Constrain first state of plan: $\bm{\tau}_k \gets \textbf{s}$
            \STATE Compute the parameters of reverse transition: $\epsilon \gets \epsilon_{\theta}(\bm{\tau}_{k},k)$, $\mu_{k-1} \gets \texttt{Denoise}(\bm{\tau}_{k},\epsilon)$
            \STATE Guide using gradients of return: $g=\nabla \mathcal{J}(\bm{\tau})|_{\bm{\tau}=\mu_{k-1}}$
            \STATE Sample: $\bm{\tau}_{k-1} \sim \mathcal{N}(\mu_{k-1}+\alpha\Sigma g,\Sigma)$
        \ENDFOR
        \STATE Execute the first action of plan $\bm{\tau}_0$
    \ENDFOR
   \end{algorithmic}
 \end{algorithm}

\paragraph{Open-loop planning.} The planner generates a trajectory (plan) $\bm{\tau}_0$ with a horizon of $T$ only at the beginning of the game. The plan is fixed and used throughout the entire episode, regardless of subsequent changes in the environment. As an illustrative example, we present the workflow of open-loop planning implemented in Diffuser~\cite{janner2022planning} in Algorithm~\ref{alg:diffuser_open_planning}.
\begin{algorithm}[h]
   \caption{Open-loop planning with Diffuser}
   \label{alg:diffuser_open_planning}
   \begin{algorithmic}[1]
    \STATE \textbf{Input:} Noise model $\epsilon_{\theta}$, guide $\mathcal{J}$
    \STATE Observe the initial state $\textbf{s}_0$
    \STATE Initialize plan $\bm{\tau}_K\sim \mathcal{N}(\bm{0},\textbf{I})$
    \FOR{$k = K \dots 1$}
        \STATE Constrain first state of plan: $\bm{\tau}_k \gets \textbf{s}_0$
        \STATE Compute the parameters of reverse transition: $\epsilon \gets \epsilon_{\theta}(\bm{\tau}_{k},k)$, $\mu_{k-1} \gets \texttt{Denoise}(\bm{\tau}_{k},\epsilon)$
        \STATE Guide using gradients of return: $g=\nabla \mathcal{J}(\bm{\tau})|_{\bm{\tau}=\mu_{k-1}}$
        \STATE Sample: $\bm{\tau}_{k-1} \sim \mathcal{N}(\mu_{k-1}+\alpha\Sigma g,\Sigma)$
    \ENDFOR
    \STATE $t=0$
    \FOR{each step of the environment}
        \STATE Execute $t$-th action of plan $\bm{\tau}_0$
        \STATE $t=t+1$
    \ENDFOR
   \end{algorithmic}
 \end{algorithm}

\newpage
\section{Proofs}\label{sec:apendix_proofs}
\textbf{Proposition~\ref{prop:probability_bound}.}~
   \emph{   
   In TAT's planning, each action is a product of aggregated information from $n$ trajectories. Consider the case of $\lambda=1$, the probability of TAT choosing the artifact can be bounded by
   \begin{equation}
      P^{\text{\rm artifact}}(n; \varUpsilon) < \frac{1}{2}\left[1-{\rm erf}\left(\frac{n/2  - n\varepsilon}{\sqrt{2 n\varepsilon(1-\varepsilon)}}\right)\right], \nonumber
   \end{equation}
   where ${\rm erf}(\cdot)$ denotes the error function.
   }

\begin{proof}
Given that the probability of artifact is $\varepsilon$, we apply the Binomial Distribution to denote the probability of having $m$ artifacts within $n$ states:
\begin{equation}
   P(m; n, \varepsilon) = \binom{n}{m} \varepsilon^m (1-\varepsilon)^{n-m},
\end{equation}
To establish an upper bound, we consider the worst-case scenario where all artifacts are the same. This means when $m$ is greater than or equal to $\lceil n/2 \rceil$, the tree will choose the artifact. Thus, we have the probability of choosing the artifact as:
\begin{equation}
   P^{\text{artifact}}(n; \varUpsilon) \leq \sum_{m=\lceil n/2 \rceil}^{n} \binom{n}{m} \varepsilon^m (1-\varepsilon)^{n-m}
\end{equation}
Next, we can use the central limit theorem to approximate this sum. Considering that $n$ is sufficiently large, the binomial distribution can be approximated by a normal distribution $\mathcal{N}(\mu, \sigma^2)$, where $\mu = n\varepsilon$ and $\sigma^2 = n\varepsilon(1-\varepsilon)$. In this case, we can express the sum as
\begin{align}
   &\int_{m=\lceil n/2 \rceil}^n \frac{1}{\sqrt{2\pi n\varepsilon(1-\varepsilon)}} \exp\left(-\frac{(m-n\varepsilon)^2}{2n\varepsilon(1-\varepsilon)}\right) \\
   < &\int_{m=n/2}^\infty \frac{1}{\sqrt{2\pi n\varepsilon(1-\varepsilon)}} \exp\left(-\frac{(m-n\varepsilon)^2}{2n\varepsilon(1-\varepsilon)}\right)\\
   = &\frac{1}{2} \left[1 - \text{erf}\left(\frac{n/2 - n\varepsilon}{\sqrt{2n\varepsilon(1-\varepsilon)}}\right)\right]\\
   = & P^{\text{upper}}(n; \varUpsilon)
\end{align}
where the error function is a mathematical function defined as $\text{erf}(z) = \frac{2}{\sqrt{\pi}} \int_{0}^{z} \exp(-t^2) dt$. 
\end{proof}

\textbf{Corollary~\ref{cor:probability_properties}.}~
\emph{For $\forall n\in \mathbb{Z}^{+}_{>1}$, the probability of TAT selecting an artifact satisfies $P^{\text{\rm artifact}}(n; \varUpsilon) < \varepsilon$ and $\lim_{n \to \infty} P^{\text{\rm artifact}}(n; \varUpsilon) = 0$.}
\begin{proof}
\textbf{(1)} We first prove $\lim_{n \to \infty} P^{\text{artifact}}(n; \varUpsilon) = 0$ as follows. 

By understanding the properties of the error function $\text{erf}(\cdot)$, we analyze its behavior as $n$ approaches infinity. Notably, the error function is monotonically increasing, and $\text{erf}(-\infty)=-1$, $\text{erf}(0)=0$, and $\text{erf}(\infty)=1$. Consequently, the upper bound of the probability $P^{\text{upper}}(n; \varUpsilon)$ approaches 0, implying that $P^{\text{artifact}}(n; \varUpsilon)$ also tends to 0. This is shown mathematically as:
\begin{equation}
   n \to \infty \Rightarrow \frac{n/2 - n\varepsilon}{\sqrt{2n\varepsilon(1-\varepsilon)}}\to \infty \Rightarrow {\rm erf}\left(\frac{n/2 - n\varepsilon}{\sqrt{2n\varepsilon(1-\varepsilon)}}\right) \to 1 \Rightarrow P^{\text{upper}}(n; \varUpsilon) \to 0 \Rightarrow P^{\text{artifact}}(n; \varUpsilon)  \to 0.
\end{equation}

\textbf{(2)} We then prove $P^{\text{artifact}}(n; \varUpsilon) < \varepsilon$ as follows.

As $\text{erf}(\cdot)$ is monotonically increasing, the upper bound probability $P^{\text{upper}}(n; \varUpsilon)$ decreases as $n$ increases:
\begin{equation}
   n \uparrow \Rightarrow P^{\text{upper}}(n; \varUpsilon) \downarrow 
\end{equation}
Thus, we have
\begin{align}
   P^{\text{upper}}(n; \varUpsilon) \leq &P^{\text{upper}}(2; \varUpsilon) = \frac{1}{2} \left[1 - \text{erf}\left(\frac{1 - 2\varepsilon}{\sqrt{4\varepsilon(1-\varepsilon)}}\right)\right].
\end{align}
We then aim to prove $P^{\text{upper}}(2; \varUpsilon)$ is a monotonically increasing convex function with respect to $\varepsilon$. This involves differentiating the function with respect to $\varepsilon$ and examining the signs of the first and second derivatives. Details are as follows.

First, we consider a composite function $F(x)$,  which is defined as a composition of two functions  $f(x)$ and $g(x)$:
\begin{equation}
   F(x)=f(g(x))=\frac{1}{2} \left[1 - \text{erf}\left(\frac{1 - 2x}{\sqrt{4x(1-x)}}\right)\right]\text{, where } f(x)=\frac{1}{2}\left[1-{\rm erf}(x)\right]\text{, } g(x)=\frac{1 - 2x}{\sqrt{4x(1-x)}}.
\end{equation}
The derivatives of $F(x)$ can be obtained by applying the chain rule:
\begin{align}
   F^{\prime}(x) &= f^{\prime}(g(x)) \cdot g^{\prime}(x), \\
   F^{\prime\prime}(x) &= f^{\prime\prime}(g(x)) \cdot g^{\prime}(x)^2 + g^{\prime\prime}(x)\cdot f^{\prime}(g(x)).
\end{align}

To proceed, we need to determine $f^{\prime}(x)$, $f^{\prime\prime}(x)$, $g^{\prime}(x)$, and $g^{\prime\prime}(x)$.

The first and second derivatives of $f(x)$ with respect to $x$ are:
\begin{align}
   &f^{\prime}(x)=\frac{\partial \frac{1}{2}\left[1-{\rm erf}(x)\right]}{\partial x} = -\frac{1}{2}\frac{\partial {\rm erf}(x)}{\partial x} = -\frac{1}{2}\frac{2}{\sqrt{\pi}} \exp(-x^2) = -\frac{1}{\sqrt{\pi}} \exp(-x^2), \\
   &f^{\prime\prime}(x)=\frac{\partial -\frac{1}{\sqrt{\pi}} \exp(-x^2)}{\partial x} = \frac{2x}{\sqrt{\pi}} \exp(-x^2).
\end{align}

For the function $g(x)$, we introduce two auxiliary functions $\varphi(x) = \sqrt{\frac{1-x}{x}}$ and $\omega(x) = \sqrt{\frac{x}{1-x}}$ for convenience. Using these, $g(x)$ can be expressed as:
\begin{align}
   g(x)=\frac{1 - 2x}{\sqrt{4x(1-x)}}=\frac{1}{2}\left(\sqrt{\frac{1-x}{x}}-\sqrt{\frac{x}{1-x}}\right)=\frac{1}{2}\left(\varphi(x)-\omega(x)\right).
\end{align}
The derivatives of auxiliary functions $\varphi(x)$ and $\omega(x)$ are easily derived:
\begin{align}
   &\frac{\partial \varphi(x)}{\partial x} = \frac{\partial \sqrt{\frac{1-x}{x}}}{\partial x} = -\frac{1}{2}\frac{\sqrt{\frac{1-x}{x}}+\sqrt{\frac{x}{1-x}}}{x}=-\frac{1}{2}\frac{\varphi(x)+\omega(x)}{x}, \\
   &\frac{\partial \omega(x)}{\partial x} =\frac{\partial \sqrt{\frac{x}{1-x}}}{\partial x} = \frac{1}{2}\frac{\sqrt{\frac{1-x}{x}}+\sqrt{\frac{x}{1-x}}}{1-x}=\frac{1}{2}\frac{\varphi(x)+\omega(x)}{1-x},
\end{align}
which are then used to find the derivatives of $g(x)$:
\begin{align}
   g^{\prime}(x)
   = &\frac{\partial \frac{1}{2}\left(\varphi(x)-\omega(x)\right)}{\partial x} \\
   = &\frac{1}{2}\left(-\frac{1}{2}\frac{\varphi(x)+\omega(x)}{x}-\frac{1}{2}\frac{\varphi(x)+\omega(x)}{1-x}\right)\\
   = &-\frac{1}{4}\left(\varphi(x)+\omega(x)\right)\left(\frac{1}{x}+\frac{1}{1-x}\right), \\
   g^{\prime\prime}(x)
   = &-\frac{1}{4}\frac{\partial \left[\left(\varphi(x)+\omega(x)\right)\left(\frac{1}{x}+\frac{1}{1-x}\right)\right]}{\partial x} \\
   = &-\frac{1}{4}\left[\frac{\partial (\varphi(x)+\omega(x))}{\partial x}\frac{1}{x(1-x)}+\frac{\partial \frac{1}{x(1-x)}}{\partial x}\Big(\varphi(x)+\omega(x)\Big)\right] \\
   = &-\frac{1}{4}\left[\left(-\frac{1}{2}\frac{\varphi(x)+\omega(x)}{x}+\frac{1}{2}\frac{\varphi(x)+\omega(x)}{1-x}\right)\frac{1}{x(1-x)}+\left(\frac{2x-1}{\left(x(1-x)\right)^2}\right)\Big(\varphi(x)+\omega(x)\Big)\right] \\
   = &-\frac{1}{8}\left[\left(\frac{2x-1}{x(1-x)}\frac{1}{x(1-x)}+\frac{2x-1}{\left(x(1-x)\right)^2}\right)\Big(\varphi(x)+\omega(x)\Big)\right] \\
   = &-\frac{1}{4}\left[\left(\frac{2x-1}{\left(x(1-x)\right)^2}\right)\Big(\varphi(x)+\omega(x)\Big)\right]. \\
\end{align}

With these derivatives in hand, we can now evaluate the first and second derivatives of $P^{\text{upper}}(2; \varUpsilon)$ with respect to $\varepsilon$ as follows:
\begin{align}
   &\frac{\partial P^{\text{upper}}(2; \varUpsilon)}{\partial \varepsilon} = F^{\prime}(\varepsilon) \\
   &= f^{\prime}(g(\varepsilon)) \cdot g^{\prime}(\varepsilon) \\
   &= \underbrace{-\frac{1}{\sqrt{\pi}} \exp\left(-\left(g(\varepsilon)\right)^2\right)}_{\text{Negative}} \cdot \underbrace{\left(-\frac{1}{4}\left(\varphi(\varepsilon)+\omega(\varepsilon)\right)\left(\frac{1}{\varepsilon}+\frac{1}{1-\varepsilon}\right)\right)}_{\text{Negative}} > 0,\\
   &\frac{\partial^2 P^{\text{upper}}(2; \varUpsilon)}{\partial \varepsilon^2} = F^{\prime\prime}(\varepsilon) \\
   &= f^{\prime\prime}(g(\varepsilon)) \cdot g^{\prime}(\varepsilon)^2 + g^{\prime\prime}(\varepsilon)\cdot f^{\prime}(g(\varepsilon)) \\
   &= \underbrace{\frac{2g(\varepsilon)}{\sqrt{\pi}} \exp\left(-\left(g(\varepsilon)\right)^2\right) \cdot g^{\prime}(\varepsilon)^2}_{\text{Positive}} + \underbrace{\left(-\frac{1}{4}\left(\frac{2\varepsilon-1}{\left(\varepsilon(1-\varepsilon)\right)^2}\right)\Big(\varphi(\varepsilon)+\omega(\varepsilon)\Big)\right)}_{\text{Negative}} \cdot \underbrace{\left(-\frac{1}{\sqrt{\pi}} \exp\left(-\left(g(\varepsilon)\right)^2\right)\right)}_{\text{Negative}} \\
   & > 0.
\end{align}
These positive first and second derivatives indicate that $P^{\text{upper}}(2; \varUpsilon)$ is indeed a monotonically increasing convex function.
Lastly, we analyze the behavior of $P^{\text{upper}}(2; \varUpsilon)$ at the extreme values of $\varepsilon$ within the range (0, 0.5):
\begin{align}
   \lim_{\varepsilon \to 0} P^{\text{upper}}(2; \varUpsilon) =  0, \ \ \ \lim_{\varepsilon \to 0.5} P^{\text{upper}}(2; \varUpsilon) =  0.5.
\end{align}
Since they are equal at the edges of (0, 0.5) and $P^{\text{upper}}(2; \varUpsilon)$ is a monotonically increasing convex function, the following inequality holds:
\begin{align}
   \forall \varepsilon \in (0, 0.5), \ \ \ \varepsilon > P^{\text{upper}}(2; \varUpsilon) \geq p^{\text{upper}}(n; \varUpsilon) > P^{\text{\rm artifact}}(n; \varUpsilon)
\end{align}
\end{proof}

\newpage
\section{Experimental Details}\label{sec:appendix_experimental_details}
\subsection{Maze2D} 
\begin{itemize}
   \item \textbf{Baseline in Maze2D.}~~
   We take the scores of MPPI, CQL, IQL, and Diffuser from Table 1 in~\cite{janner2022planning}. 
   \item \textbf{Tree planners in Maze2D.}~~
   We deploy TAT on Diffuser, using the pre-trained models of Diffuser released from the authors: (\emph{Diffuser Repository}, \url{https://github.com/anuragajay/decision-diffuser}). We adopt default hyperparameters from Diffuser's implementation. Regarding the additional hyperparameters of TAT, we set $\lambda=0.98$ and $1-\alpha=0.0005$. The bath size of sampled trajectories is set to $128$.
\end{itemize}

\subsection{Block Stacking} 

\begin{itemize}
   \item \textbf{Baseline in block stacking.}~~
   We take the scores of BCQ, and CQL from Table 2 in~\cite{janner2022planning}. We ran Diffuser using the official implementation and released pre-trained models from \emph{Diffuser Repository}. Given that there are no publicly available pre-trained models of Decision Diffuser (DD), we retrain DD using the official implementation and default hyperparameters from the authors: (\emph{DD Repository}, \url{https://github.com/anuragajay/decision-diffuser}). 
   \item \textbf{Tree planners in block stacking.}~~
   $\text{Diffuser}^{\varUpsilon}$ uses the pre-trained models of Diffuser released from \emph{Diffuser Repository} and inherits all default hyperparameters and settings. $\text{DD}^{\varUpsilon}$ is deployed on top of DD's official implementation and inherits all default hyperparameters and settings from DD. Regarding the hyperparameters of TAT, we set $\lambda=0.98$ and $1-\alpha=0.002$. The bath size of trajectories is set to $64$.
\end{itemize}

\subsection{Locomotion} 

\begin{itemize}
   \item \textbf{Baseline in locomotion.}~~
   We take the scores of BC, CQL, and IQL from Table 1 in~\cite{kostrikov2022offline}; DT from Table 2 in~\cite{chen2021decision}; TT from Table 1 in~\cite{janner2021offline}; MOPO from Table 1 in~\cite{yu2020mopo}; MOReL from Table 2 in~\cite{kidambi2020morel}; MBOP from Table 1 in~\cite{argenson2021modelbased}; Diffuser from Table 2 in~\cite{janner2022planning}; RGG from Table 1 in~\cite{lee2023refining}.

   \item \textbf{Tree planners in locomotion.}~~
   Both $\text{Diffuser}^{\varUpsilon}$ and $\text{RGG}^{\varUpsilon}$ are deployed on the pre-trained models of Diffuser and Restoration Gap Guidance (RGG), available in their respective official repositories \emph{Diffuser Repository} and \emph{RGG Repository} (\url{https://github.com/leekwoon/rgg}). $\text{Diffuser}^{\varUpsilon}$ and $\text{RGG}^{\varUpsilon}$ keep all default hyperparameters from Diffuser and RGG. For the hyperparameters of TAT, we set $\lambda=0.98$, $1-\alpha=0.005$ for Walker2d and Hopper, and $1-\alpha=0.002$ for HalfCheetah. We found that we could reduce the sampling step for many tasks through $\text{warm-start}^{\varUpsilon}$ planning (e.g., from default 20 to 10 in the Hopper tasks). Since Diffuser and RGG generate default $64$ trajectories for each planning step, we use the first $32$ (half) for TAT construction.
\end{itemize}

\subsection{Other Details} 
\begin{itemize}
   \item Different diffusion planners have slight variations in their network architectures for diffusion models, but overall, they all feature a U-Net architecture~\cite{ronneberger2015u} with repeated convolutional residual blocks. Each residual block consists of two convolutional layers, followed by group norm~\cite{wu2018group} and a Mish activation function~\cite{misra2019mish}.
   \item All experiments are run on the NVIDIA GeForce RTX 3080 GPU core.
\end{itemize}

\newpage
\section{The Impact of the Aggregated Trajectory Number}
Proposition~\ref{prop:probability_bound} previously established that the likelihood of artifact selection decreases as the number $n$ of involved trajectories increases. It suggests that a higher number leads to more stable and trustworthy outcomes. To validate this, we varied the batch size for $\text{Diffuser}^{\varUpsilon}$ in the Hopper Medium task and measured the average number of trajectories contributing to an action generation. The results are shown in Table~\ref{tab:appendix_number}. We can see that more trajectories correlate with better performance, which is consistent with the theoretical analysis. However, a higher number also means a greater computational budget. Meanwhile, the performance gain becomes marginal as the number becomes sufficiently large. Thus, there is a trade-off.
\begin{table}[h]
   \caption{The performance of $\text{Diffuser}^{\varUpsilon}$ under varying numbers of trajectories in the Hopper Medium task. }
   \vskip 0.1in
   \label{tab:appendix_number}
   \begin{center}
   \begin{small}
   % \begin{sc}
   \begin{tabular}{rcc}
         \toprule
         \textbf{Batch Size} & \textbf{Average Trajectories per Action} & \textbf{Score}  \\
         \midrule
         \phantom{0}\phantom{0}1        & \phantom{0}\phantom{0}1.0{\tiny$\pm$0.0}             & 74.0{\tiny$\pm$2.2}      \\
         \phantom{0}\phantom{0}8        & \phantom{0}13.4{\tiny$\pm$0.3}            & 77.4{\tiny$\pm$2.4}       \\
         \phantom{0}16       & \phantom{0}24.0{\tiny$\pm$0.4}            & 80.7{\tiny$\pm$2.3}       \\
         (Default)\phantom{0}32       & \phantom{0}49.4{\tiny$\pm$0.7}           & 82.6{\tiny$\pm$1.9}       \\
         \phantom{0}64       & \phantom{0}96.1{\tiny$\pm$0.9}           & 84.5{\tiny$\pm$2.0}     \\
         128      & 179.9{\tiny$\pm$1.2}             & 85.2{\tiny$\pm$2.0}    \\
         256      & 358.7{\tiny$\pm$3.0}           & 86.2{\tiny$\pm$2.1}    \\
         512      & 668.1{\tiny$\pm$6.5}            & 86.9{\tiny$\pm$1.9}    \\
         \bottomrule
   \end{tabular}
   % \end{sc}
   \end{small}
   \end{center}
   \vskip -0.1in
\end{table}

\newpage
\section{Time and Memory Budget}\label{sec:planning_time}
In this part, we analyze the additional time and memory budget for maintaining TAT. 

We present the average planning time of generating an action taken by baseline Diffuser and $\text{Diffuser}^{\varUpsilon}$ on MuJoCo locomotion~\cite{fu2020d4rl} in Table~\ref{tab:planning_time}. Notably, the time consumption of TAT is less than $5\%$ of the total planning time. 

We present the memory usage of baseline Diffuser and $\text{Diffuser}^{\varUpsilon}$ on Hopper Medium task in Table~\ref{tab:memory_usage}. As shown, the memory usage for the tree is small. TAT typically manages an average of 200-400 nodes, with each node only storing 3 statistics. This volume of data is significantly smaller than that of U-Net’s sampling in the vanilla diffusion planner.

\begin{table}[h]
   \caption{The planning time of $\text{Diffuser}^{\varUpsilon}$ and Diffuser on MuJoCo locomotion tasks. The results are averaged over 200 actions' generation. All results are tested on a single NVIDIA GeForce RTX 3080 GPU core. The unit is second (s).}
   \label{tab:planning_time}
   \vskip 0.1in
   \begin{center}
   \begin{small}
   % \begin{sc}
   \begin{tabular}{llccc}
         \toprule
         \textbf{*-Medium} & \textbf{Method} & \textbf{Sampling Time}  & \textbf{Tree  Time} & \textbf{Total} \\
         \midrule
         HalfCheetah  & Diffuser                          & 1.43  & -      & 1.43     \\
         HalfCheetah  & $\text{Diffuser}^{\varUpsilon}$   & 1.43  & 0.02   & 1.45     \\
         \midrule
         Hopper      &  Diffuser                         & 1.68  & -   & 1.68   \\
         Hopper      &  $\text{Diffuser}^{\varUpsilon}$  & 1.68  & 0.06   & 1.74   \\
         \midrule
         Walker2d  & Diffuser                          & 1.69  & -      & 1.69     \\
         Walker2d  & $\text{Diffuser}^{\varUpsilon}$   & 1.68  & 0.07   & 1.75     \\
         \bottomrule
   \end{tabular}
   % \end{sc}
   \end{small}
   \end{center}
   \vskip -0.1in
\end{table}

\begin{table}[h]
   \caption{The memory usage of $\text{Diffuser}^{\varUpsilon}$ and Diffuse on Hopper Medium task. All results are tested on a single NVIDIA GeForce RTX 3080 GPU core.}
   \label{tab:memory_usage}
   \vskip 0.1in
   \begin{center}
   \begin{small}
   % \begin{sc}
   \begin{tabular}{lcccc}
         \toprule
          \textbf{Method} & \textbf{U-Net Sampling Memory}  & \textbf{Tree Memory} & \textbf{Other Fixed Memory} & \textbf{Total} \\
         \midrule
         Diffuser  & 2501.7 MiB  & -      & 2965.3 MiB& 5467.1 MiB\\
          $\text{Diffuser}^{\varUpsilon}$   & 2501.9 MiB & 16.8 MiB  & 2964.9 MiB & 5483.7 MiB  \\
         \bottomrule
   \end{tabular}
   % \end{sc}
   \end{small}
   \end{center}
   \vskip -0.1in
\end{table}

\newpage
\section{Visual Comparison in Maze2D}
\begin{figure*}[h]
   \vskip 0.1in
   \begin{center}
   \centerline{\includegraphics[width=0.9\textwidth]{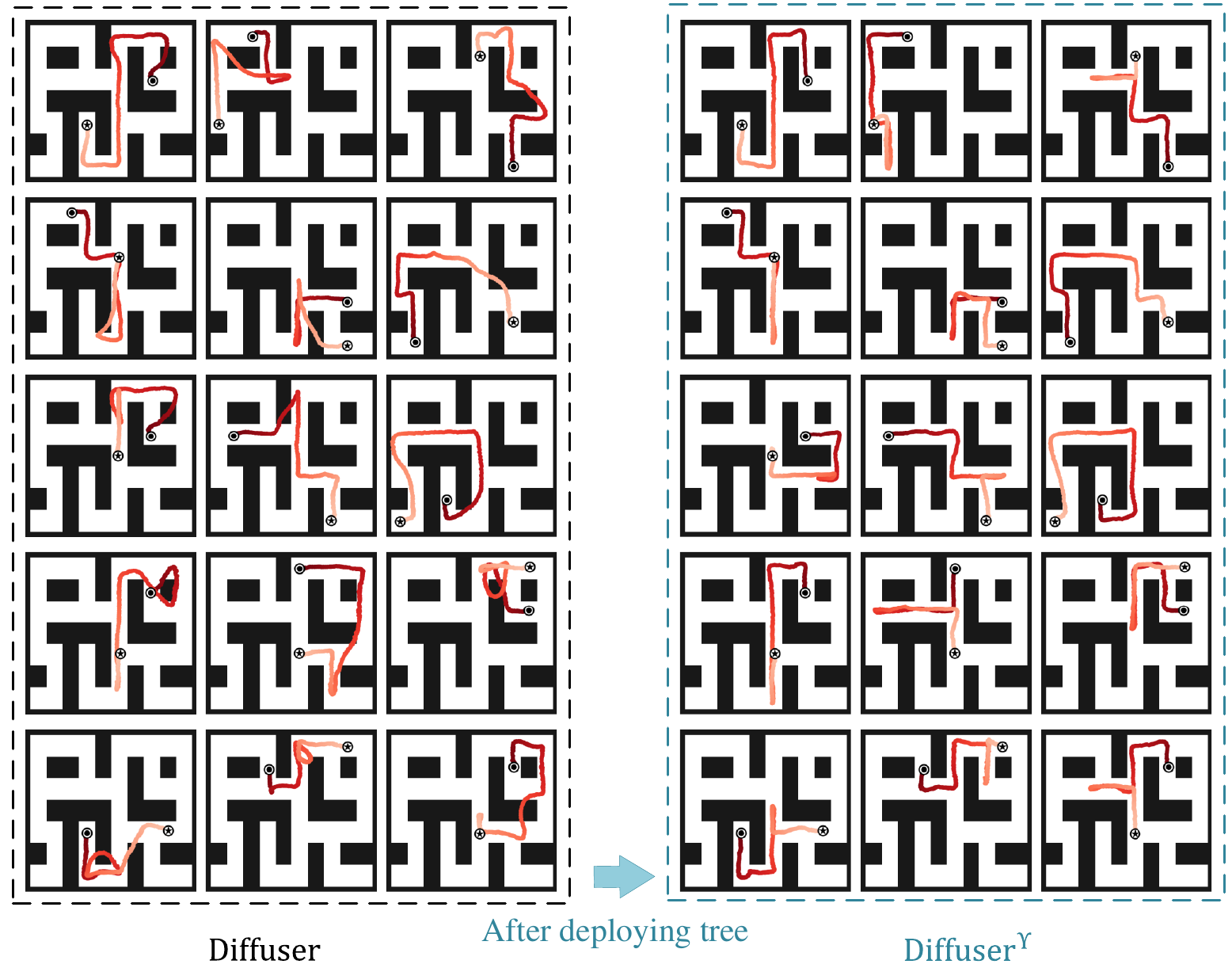}}
   \caption{Visual comparison of Diffuser and $\text{Diffuser}^{\varUpsilon}$ in the presence of artifacts in the Maze2D-Large environments. \protect{\raisebox{-.05cm}{\includegraphics[height=.3cm]{figures/mark_start_crop.png}}} denotes the starting position and \protect{\raisebox{-.05cm}{\includegraphics[height=.3cm]{figures/mark_goal_crop.png}}} denotes the goal position.}
   \label{fig:appendix_maze_results}
   \end{center}
   \vskip -0.2in
\end{figure*}

We show the stochastic infeasible plans of Diffuser in the Maze2D environments in Figure~\ref{fig:appendix_maze_results}. Given the same scenarios, $\text{Diffuser}^{\varUpsilon}$ can effectively filter them out by TAT. 

\newpage
\section{Visual Comparison in Kuka Block Stacking}
\begin{figure*}[h]
   \vskip -0.1in
   \begin{center}
   \centerline{\includegraphics[width=0.85\textwidth]{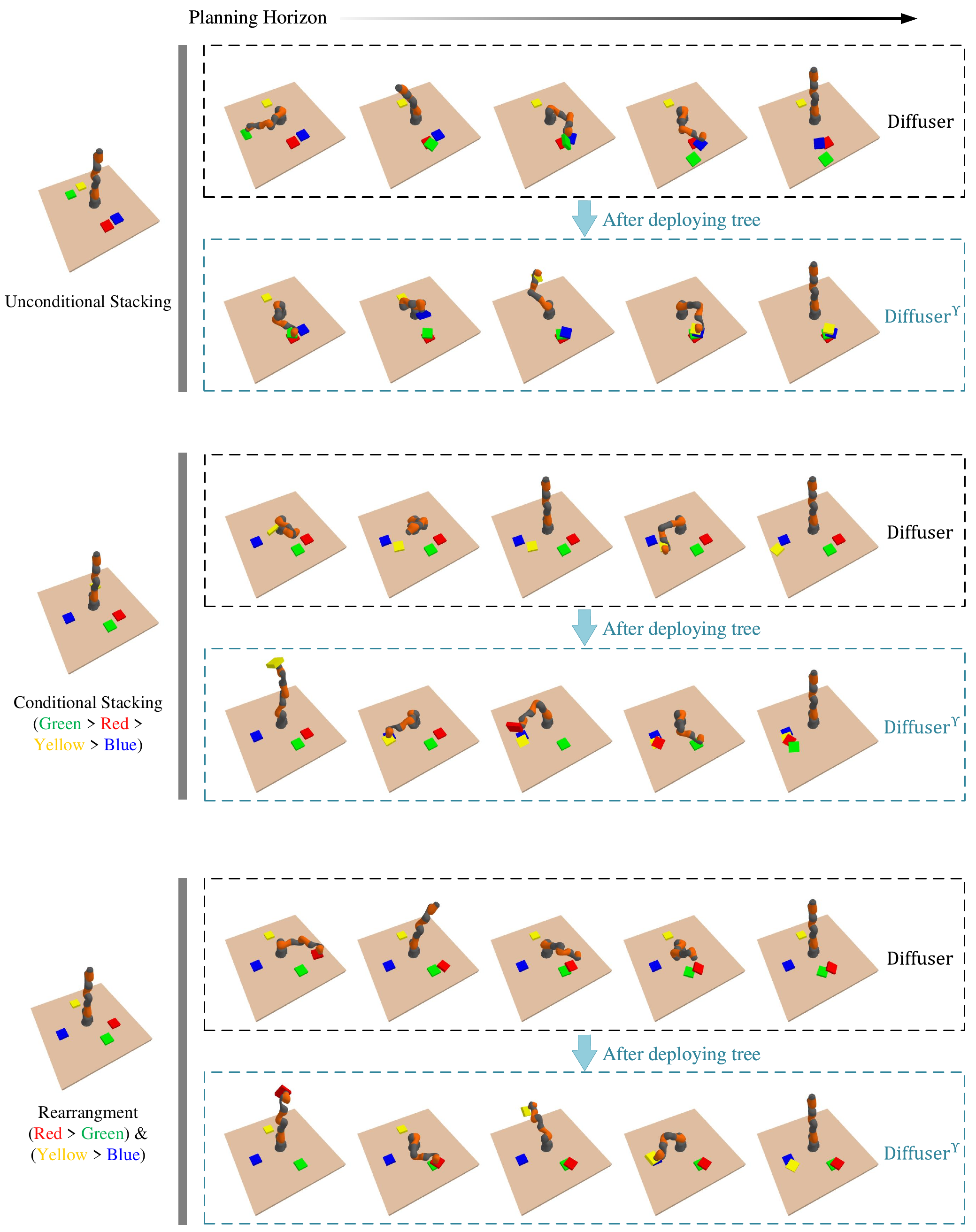}}
   \caption{Visual comparison of Diffuser and $\text{Diffuser}^{\varUpsilon}$ in the presence of artifacts in the Kuka block stacking environments. }
   \label{fig:appendix_kuka_results}
   \end{center}
   \vskip -0.8in
\end{figure*}

We present the stochastic artifacts of Diffuser in the Kuka block stacking environments in Figure~\ref{fig:appendix_kuka_results}. Given the same scenarios, $\text{Diffuser}^{\varUpsilon}$ can effectively handle these artifacts by TAT. 

\newpage
\section{Visual Comparison in MuJoCo Locomotion}
\begin{figure*}[h]
   \begin{center}
   \centerline{\includegraphics[width=0.87\textwidth]{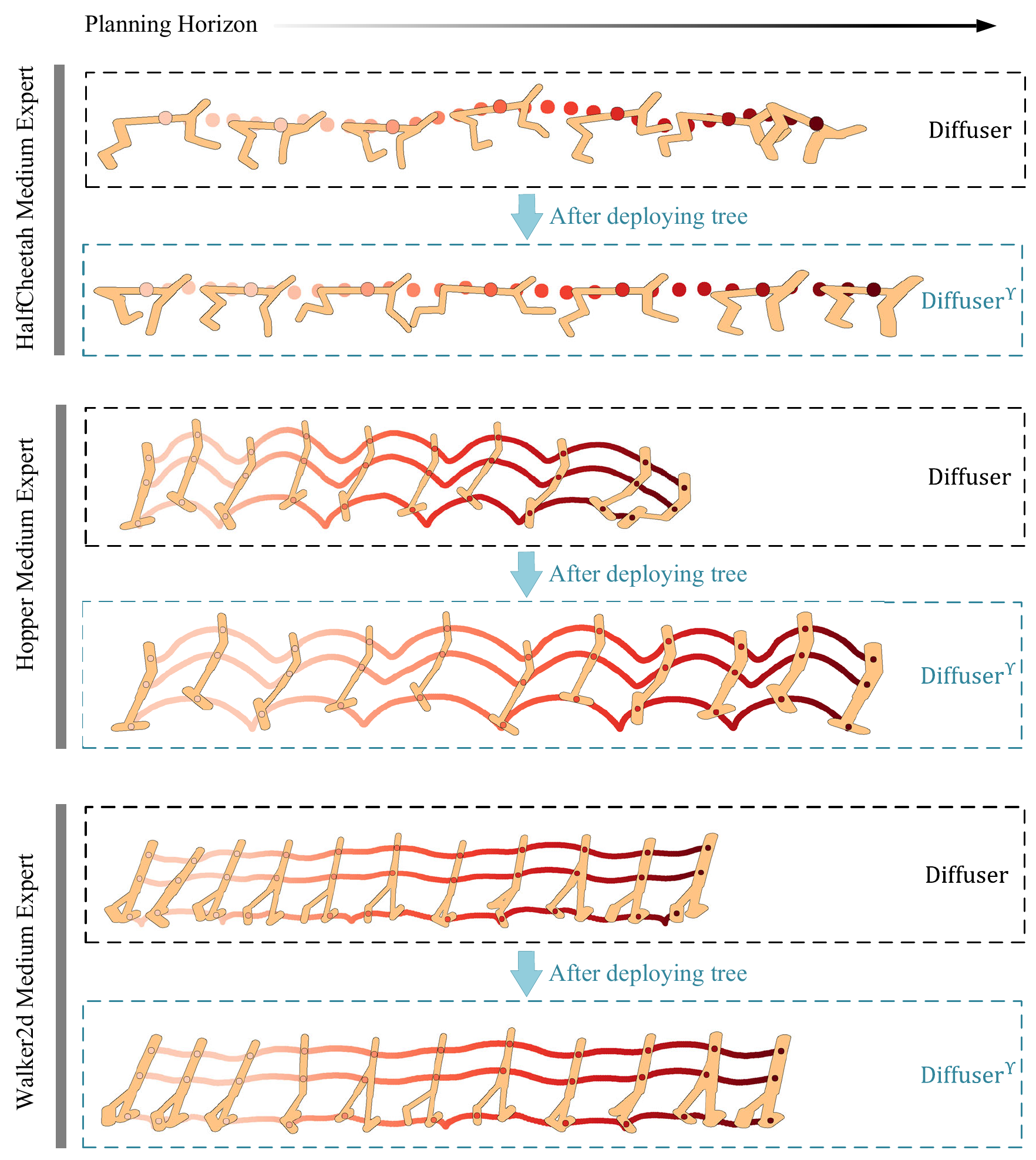}}
   \caption{Visual comparison of Diffuser and $\text{Diffuser}^{\varUpsilon}$ in the presence of artifacts in the MuJoCo locomotion tasks.}
   \label{fig:appendix_mujoco_results}
   \end{center}
   \vskip -0.2in
\end{figure*}

We show the stochastic suboptimal and unreliable plans of Diffuser in the MuJoCo locomotion tasks in Figures~\ref{fig:appendix_mujoco_results}. Given the same scenarios, $\text{Diffuser}^{\varUpsilon}$ can effectively handle these suboptimal and unreliable plans by TAT. 

\newpage
\section{Artifacts in Vision Domains}
The artifacts, denoting defective or infeasible outcomes, have been a key concern in studies on generative neural networks, particularly in vision domains. \citet{bau2019gan,Zhang2019detecting,tousi2021automatic} have made efforts to understand the inner workings of Generative Adversarial Networks (GAN). These studies identified and adjusted the internal units that produced artifacts within GANs. \citet{shen2020interpreting} further elucidated how the latent space coded different semantics and trained a linear classifier based on artifact-labeled data to mitigate artifact generation. Other works~\cite{choi2022can,jeong2022unsupervised} have been centered on unsupervised methods for detecting and rectifying artifacts in generative models.

\end{document}